%% file: el-pv-cell-segmentation.tex
\pdfoutput=1
\let\arxiv\empty
\RequirePackage{fix-cm}
\PassOptionsToPackage{american}{babel}
\documentclass[twocolumn]{svjour3}
\usepackage{iftex}
\ifLuaTeX
\usepackage{luatexbase}
\fi
\usepackage[babel]{microtype}
\usepackage[T1]{fontenc}
\usepackage[utf8]{inputenc}
\usepackage[inline]{enumitem}
\usepackage{caption}
\usepackage{subcaption}
\usepackage{rebuttal}
\usepackage{amsmath}
\usepackage{amssymb}
\usepackage{babel}
\usepackage{mathtools}
\usepackage[babel,style=american,autopunct]{csquotes}
\usepackage[binary-units,detect-all,per-mode=symbol,range-units=single]{siunitx}
\usepackage[l2tabu,orthodox]{nag}
\usepackage{tikz}
\usepackage{tikzscale}
\usepackage{tikzscalefix}
\usepackage{tikznonag}
\usepackage{dblfloatfix}
\usepackage{pgfplots}
\usepackage{pgfplotstable}
\usepackage{booktabs}
\usepackage{csquotes}
\usepackage{xcolor}
\usepackage{multirow}
\usepackage{multicol}
\usepackage{xfrac}
\usepackage{tabularx}
\usepackage{etoolbox}
\usepackage[acronym]{glossaries}
\usepackage[all]{nowidow}
\usepackage[numbers]{natbib}
\usepackage[abbreviations]{foreign}
\usepackage[switch,mathlines]{linenumbers}
\usepackage{hyperref}
\usepackage[hyperpageref]{backref}
\usepackage[a4paper,includehead,includefoot,margin=0.75in]{geometry}

\setlist[description]{font=\normalfont\bfseries}

\makeatletter
\def\cl@chapter{\@elt {theorem}}
\makeatother

\usepackage[capitalise]{cleveref}

\hypersetup{bookmarksdepth=4}

\pgfplotsset{compat=1.13}

\usetikzlibrary{angles}
\usetikzlibrary{arrows.meta}
\usetikzlibrary{arrows}
\usetikzlibrary{backgrounds}
\usetikzlibrary{calc}
\usetikzlibrary{decorations}
\usetikzlibrary{external}
\usetikzlibrary{fit}
\usetikzlibrary{intersections}
\usetikzlibrary{matrix}
\usetikzlibrary{patterns}
\usetikzlibrary{positioning}
\usetikzlibrary{spy}
\usetikzlibrary{topaths}

\tikzset{>=latex'}

\usepgfplotslibrary{colorbrewer}
\usepgfplotslibrary{colormaps}
\usepgfplotslibrary{fillbetween}
\usepgfplotslibrary{groupplots}
\usepgfplotslibrary{statistics}
\usepgfplotslibrary{units}

\newcommand\bitwise[1]{%
  bitwise{\footnotesize~\textsf{\MakeUppercase{#1}}}%
}

\newcommand\cplus{\begingroup\small+\endgroup}
\newcommand*{\basecpp}{C\raisebox{.2ex}{\textsf{\kern-.15ex\cplus\kern-.3ex\cplus}}}
\newcommand*{\cpp}{\basecpp} 

\newcommand\proj{\vec\pi}

\newcommand\transp{\top\kern-\scriptspace}
\newcommand\distort{\vec{\delta}}
\newcommand\undistort{\distort^{-1}}
\newcommand\normcoord{\vec n}

\hypersetup{
  colorlinks,
  pdflang=en-US,
  pdfpagemode=UseNone,
  pdfpicktraybypdfsize=true,
  pdfprintscaling=None,
  pdfstartview=FitH,
  pdftitle={Segmentation of Photovoltaic Module Cells in Uncalibrated Electroluminescence
  Images},
  pdfauthor={Sergiu Deitsch},
  pdfsubject={Instance segmentation},
  pdfkeywords={PV modules, EL imaging, visual inspection, lens distortion, solar
  cell extraction, pixelwise classification},
}

\makeatletter

\newcommand{\doi}[1]{%
  \begingroup
    \let\bibinfo\@secondoftwo
    \textsc{doi}:\,%
    \discretionary{}{}{}%
    \href{http://dx.doi.org/#1}{%
      \nolinkurl{#1}%
    }%
  \endgroup
}

\makeatother

\let\mat\tens

\newcolumntype{Y}{>{\centering\arraybackslash}X}

\DeclareMathOperator*{\argmin}{argmin}

\DeclareMathOperator\diag{diag}

\DeclarePairedDelimiter\norm{\lVert}{\rVert}

\glsdisablehyper

\newacronym{AP}{AP}{Average Precision}
\newacronym{AUC}{AUC}{Area Under the Curve}
\newacronym{CCD}{CCD}{Charge-coupled Device}
\newacronym{CGS}{CGS}{Curve Grid Segmentation}
\newacronym{CNN}{CNN}{Convolutional Neural Network}
\newacronym{CPD}{CPD}{Coherent Point Drift}
\newacronym{CRF}{CRF}{Conditional Random Field}
\newacronym{DLT}{DLT}{Direct Linear Transform}
\newacronym{EL}{EL}{electroluminescence}
\newacronym{Fast R-CNN}{Fast R-CNN}{Fast Region-based Convolutional Neural Network}
\newacronym{FCIS}{FCIS}{Fully Convolutional Instance Segmentation}
\newacronym{FCN}{FCN}{Fully Convolutional Network}
\newacronym{FOV}{FOV}{Field-of-View}
\newacronym{GN}{GN}{Gauss-Newton}
\newacronym{IoC}{IoC}{Intersection-over-Union}
\newacronym{IoU}{IoU}{Intersection-over-Union}
\newacronym{IQR}{IQR}{interquartile range}
\newacronym{IR}{IR}{infrared}
\newacronym{IV}{I-V}{current-voltage characteristic}
\newacronym{LED}{LED}{Light-emitting Device}
\newacronym{LO-RANSAC}{LO-RANSAC}{Locally Optimized RANdom SAmple Consensus}
\newacronym{MLS}{MLS}{Moving Least Squares}
\newacronym{MSE}{MSE}{Mean Squared Error}
\newacronym{PGA}{PGA}{Perspective-corrected Grid Alignment}
\newacronym{PR}{PR}{Precision-Recall}
\newacronym{PV}{PV}{photovoltaic}
\newacronym{R-CNN}{R-CNN}{Regions with CNN features}
\newacronym{RANSAC}{RANSAC}{RANdom SAmple Consensus}
\newacronym{RMSE}{RMSE}{Root Mean Square Error}
\newacronym{RPN}{RPN}{Region Proposal Network}
\newacronym{SVD}{SVD}{Singular Value Decomposition}
\newacronym{SVM}{SVM}{Support Vector Machine}
\newacronym{TPE}{TPE}{Tree-Structured Parzen Estimator}

\input{pgfstyles}

\tikzset{
  external pdflatex/.style={external/system call={pdflatex \tikzexternalcheckshellescape -halt-on-error -interaction=batchmode -jobname "\image" "\texsource"}},
  external pdflatex,
}

\tikzset{external/mode=list and make}

\tikzifexternalizing{%
  \renewcommand\pgfsyspdfmark[3]{}%
}{}

\begin{document}

\newcommand\cellmarker[1][major grid intersections only]{%
  \tikzsetnextfilename{cell-marker}%
  \tikz [/pgfplots/every crossref picture,inner sep=2pt] {%
    \pgfplotsset{line legend,
      legend image code={thick,color=major grid,#1},
      enlarge legend image=0.5*\pgfplotmarksize,
    }%
  }%
}

\newcommand\busbarmarker[1][minor grid intersections only]{%
  \tikzsetnextfilename{busbar-marker}%
  \tikz [/pgfplots/every crossref picture] {%
    \pgfplotsset{line legend,
      legend image code={thick,densely dashed,color=busbars,#1},
      enlarge legend image=0.5*\pgfplotmarksize,
    }%
  }%
}

\newcommand\linemarker[1]{%
  \tikz [/pgfplots/every crossref picture] {%
    \pgfplotsset{line legend,legend image code={#1}}%
  }%
}

\newcommand\gridmarker{%
  \tikzsetnextfilename{grid-marker}%
  \linemarker{thick,Firebrick2}%
}

\newcommand\vcurvemarker{%
  \tikzsetnextfilename{vertical-curve-marker}%
  \linemarker{parabola/vertical}%
}

\newcommand\hcurvemarker{%
  \tikzsetnextfilename{horizontal-curve-marker}%
  \linemarker{parabola/horizontal}%
}

\sisetup{
  round-mode=places,
  round-precision=2,
  zero-decimal-to-integer,
  group-minimum-digits=3,
  group-separator={{,}},
}

\pgfkeys{
  /pgf/number format/.cd,
  fixed,
  precision=2,
}

\normalem

\ifx\arxiv\undefined\else
\rebuttalset{marked=false}
\fi


\ifx\arxiv\undefined
\rebuttalset{deletion/command=\sout}
\input{rebuttal2}
\fi

\glsresetall

\title{Segmentation of Photovoltaic Module Cells in Uncalibrated Electroluminescence Images}
\author{%
  Sergiu Deitsch \and
  Claudia Buerhop-Lutz \and
  Evgenii Sovetkin \and
  Ansgar Steland \and
  Andreas Maier \and
  Florian Gallwitz \and
  Christian Riess
}

\institute{%
  S.~Deitsch \at
  Pattern Recognition Lab\\
  University of Erlangen-Nuremberg\\
  Martensstr.~3\\
  91058~Erlangen, Germany\\
  \email{\href{mailto:sergiu.deitsch@fau.de}{sergiu.deitsch@fau.de}}
}

\date{Received: date / Accepted: date}


\maketitle

\ifx\arxiv\undefined
\linenumbers
\fi

\begin{abstract}
High resolution \gls{EL} images captured in the infrared spectrum allow to
visually and non-destructively inspect the quality of \gls{PV} modules.
Currently, however, such a visual inspection requires trained experts to discern
different kinds of defects, which is time-consuming and expensive. Automated
segmentation of cells is therefore a key step in automating the visual
inspection workflow.

In this work, we propose a robust automated segmentation method for extraction
of individual solar cells from \gls{EL} images of \gls{PV} modules. This enables
controlled studies on large amounts of data to understanding the effects of
module degradation over time---a process not yet fully understood.

The proposed method infers in several steps a high-level solar module
representation from low-level ridge edge features.  An important step in the
algorithm is to formulate the segmentation problem in terms of lens calibration
by exploiting the plumbline constraint. We evaluate our method on a dataset of
various solar modules types containing a total of~\num{408}~solar cells with
various defects. Our method robustly solves this task with a median weighted
Jaccard index of~\SI{94.46674192878733}{\percent} and an~\(F_1\)~score of
\SI{97.61704495045228}{\percent}, both indicating
\addition[label=a:robustness,ref=c:contrib2]{a high sensitivity and} a high
similarity between automatically segmented and ground truth solar cell masks.

\end{abstract}

\keywords{
  \Gls{PV} modules,
  \gls{EL} imaging,
  visual inspection,
  lens distortion,
  solar cell extraction,
  pixelwise classification
}


\section{Introduction}
\label{sec:intro}

Visual inspection of solar modules using \gls{EL} imaging allows to easily
identify damage inflicted to solar panels either by environmental influences
such as hail, during the assembly process, or due to prior material defects or
material
aging~\cite{Nian2010,Breitenstein2011,Tsai2012,Tsai2013,Anwar2014,Tseng2015}.
The resulting defects can notably decrease the photoelectric conversion
efficiency of the modules and thus their energy yield. This can be avoided by
continuous inspection of solar modules and maintenance of defective units.
For an introduction and review of non-automatic processing tools for \gls{EL}
images, we refer to~\citet{Mauk2013}.

An important step towards an automated visual inspection is the segmentation of
individual cells from the solar module. An accurate segmentation allows to
extract spatially normalized solar cell images.
\change[label=ch:next-steps,ref=c:pv-advances2]{Such solar cell images are the
  ideal training data for classifiers to predict defects in solar
  modules~\cite{Deitsch2019}} {We already used the proposed method to develop a
  public dataset of solar cells images~\cite{Buerhop2018}, which are highly
  accurate training data for classifiers to predict defects in solar
  modules~\cite{Deitsch2019,Mayr2019}}. In particular,
  \change[label=ch:samples-aligned,ref=c:cnn-invariance]{samples fed into a
    \gls{CNN} need to be correctly aligned}{the \gls{CNN} training is greatly simplified when using spatially normalized samples,}
    because \glspl{CNN} are
    \addition[label=a:cnn-equiv,ref=c:cnn-invariance]{generally able to learn
      representations that are only equivariant to small
      translations~\cite[pp.~335--336]{Goodfellow2016}.}
      \change[label=ch:cnn-spatial,ref=c:cnn-invariance]{but in general not
        spatially invariant~\hbox{\cite{Jaderberg2015,Lin2017}}}{The learned
      representations, however, are not naturally invariant to other spatial
    deformations such as rotation and
  scaling~\cite{Goodfellow2016,Jaderberg2015,Lin2017}}.

The identification of solar cells is additionally required by the international
technical specification IEC TS~60904-13~\cite[Annex~D]{IEC2018} for further
identification of defects on cell level. Automated segmentation can also ease
the development of models that predict the performance of a \gls{PV} module
based on detected or identified failure modes, or by determining the operating
voltage of each cell~\cite{Potthoff2010}. The data describing the cell
characteristics can be fed into an electric equivalent model that allows to
estimate or simulate the \gls{IV}
curve~\cite{Quaschning1996,Chenni2007,Karatepe2007} or even the overall power
output~\cite{Kaushika2003}.


\begin{figure*}[tb]
  \centering
  \tikzsetnextfilename{cell-extraction-overview}%
  \input{figures/cell-extraction-overview}%
  \caption[]{\subref{fig:el-pv-curve-grid}~An \gls{EL} image of a \gls{PV}
    module overlaid by a rectangular grid~(\gridmarker) and parabolic curve
    grid~(\cellmarker) including the busbars~(\busbarmarker) determined using
    our approach. The intersections of the rectangular grid were registered to
    curve grid intersections to accurately align both grids. Notice how the
    rectangular grid is still not able to capture the curved surface of the
    solar module induced by the (weak) lens distortion that increases especially
    towards the image border. Using the curve grid, we estimate the lens
    distortion, rectify the image and finally extract the individual cells using
    the estimated module topology~\subref{fig:el-pv-extracted-cells}. The
    segmented solar cells can be used for further analysis, such as automatic
    defect classification or failure prediction in \gls{PV} modules. The solar
    cells are approximately~\(\SI{15.6}{\centi\metre} \times
    \SI{15.6}{\centi\metre}\) with a standard 60~cell \gls{PV} module with
    overall dimensions of \(\SI{1}{\metre}\times \SI{1.65}{\metre}\).}
  \label{fig:el-pv-image-to-cells}
\end{figure*}

The appearance of \gls{PV} modules in \gls{EL} images depends on a number of
different factors, which makes an automated segmentation challenging. The
appearance varies with the type of semiconducting material and with the shape
of individual solar cell wafers. Also, cell cracks and other defects can
introduce distracting streaks. A solar cell completely disconnected from the
electrical circuit will also appear much darker than a functional cell.
Additionally, solar modules vary in the number of solar cells and their layout,
and solar cells themselves are oftentimes subdivided by busbars into multiple
segments of different sizes. Therefore, it is desirable for a fully automated
segmentation to infer both the arrangement of solar cells within the \gls{PV}
module and their subdivision from \gls{EL} images alone, in a way that is
robust to various disturbances. In particular, this may ease the inspection of
heterogeneous batches of \gls{PV} modules.

\phantomsection
\label{sec:manufacturing-setting}

In this work, we assume that \gls{EL} images are captured in a manufacturing
setting \addition[label=a:setting,ref=c:setting]{or under comparable
conditions in a test laboratory where field-aged modules are analyzed either
regularly or after hazards like hailstorms. Such laboratories oftentimes
require agile work processes where the equipment is frequently remounted}. In
these scenarios, the \gls{EL} irradiation of the solar module predominates the
background
irradiation, and the solar modules are captured facing the \gls{EL} camera
without major perspective distortion. Thus, the geometric distortions that are
corrected by the proposed method are radial lens distortion, in-plane rotation,
and minor perspective distortions. This distinguishes the manufacturing setting
from acquisitions in the field, where \gls{PV} modules may be occluded by cables
and parts of the rack, and the perspective may be strong enough to require
careful correction. However, perspective distortion also makes it more difficult
to identify defective areas (\eg, microcracks) due to the foreshortening
effect~\cite{Aloimonos1988}. Therefore, capturing \gls{EL} images from an
extreme perspective is generally not advisable. Specifically for manufacturing
environments, however, the proposed method yields a robust, highly accurate, and
completely automatic segmentation of solar modules into solar cells from high
resolution \gls{EL} images of \gls{PV} modules.

\addition[label=a:setting-goal,ref=c:setting2]{Independently of the setting, our
goal is to allow for some flexibility for the user to freely position the
camera or use zoom lenses without the need to recalibrate the camera.}

\addition[label=a:calibration-pattern,ref=c:contrib2]{With this goal in mind, a
particular characteristic of the proposed segmentation pipeline is that it
does not require an external calibration pattern. During the detection of the
grid that identifies individual solar cells, the busbars and the inter solar
cell borders are directly used to estimate lens distortion. Avoiding the use of
a separate calibration pattern also avoids the risk of an operator error during
the calibration, \eg, due to inexperienced personnel.}

A robust and fully automatic \gls{PV} module segmentation can help understanding
the influence of module degradation on module efficiency and power generation.
Specifically, this allows to continuously and automatically monitor the
degradation process, for instance, by observing the differences in a series of
solar cell images captured over a certain period of time. The segmentation also
allows to automatically create training data for learning-based algorithms for
defect classification and failure prediction.

\subsection{Contributions}
\label{sec:contributions}

To the best of our knowledge, the proposed segmentation pipeline is the first
work to enable a fully automatic extraction of solar cells from uncalibrated
\gls{EL} images of solar modules (\cf, \cref{fig:el-pv-extracted-cells}).
\addition[label=a:goal,ref=c:training]{Within the pipeline, we seek to obtain
the exact segmentation mask of each solar cell through estimation of
non-linear and linear transformations that warp the \gls{EL} image into a
canonical view.} \addition[label=a:contributions,ref=c:training] {To this end,
our contributions are three-fold:
\begin{enumerate}
  \item Joint camera lens distortion estimation and \gls{PV} module grid
    detection for precise solar cell region identification.
  \item A robust initialization scheme for the employed lens distortion model.
  \item A highly accurate pixelwise classification into active solar cell area
    on monocrystalline and polycrystalline \gls{PV} modules robust to various
    typical defects in solar modules.
\end{enumerate}
Moreover, our method operates on arbitrary (unseen) module layouts without prior knowledge on the layout.}


\subsection{Outline}

The remainder of this work is organized as follows.
\Cref{sec:el-pv-seg-related-work} discusses the related work. In
\cref{sec:el-pv-seg-methodology}, the individual stages of the segmentation
pipeline are presented. In \cref{sec:el-pv-seg-evaluation}, we evaluate the
presented segmentation approach on a number of different \gls{PV} modules with
respect to the segmentation accuracy. Finally, the conclusions are given in
\cref{sec:el-pv-seg-conclusions}.

\section{Related Work}
\label{sec:el-pv-seg-related-work}

The segmentation of \gls{PV} modules into individual solar cells is related to
the detection of calibration patterns, such as checkerboard patterns commonly
used for calibrating intrinsic camera and lens
parameters~\cite{Rufli2008,Placht2014, Fuersattel2016,Hoffmann2017, Ha2017}.
However, the appearance of calibration patterns is typically perfectly known,
whereas detection of solar cells is encumbered by various defects that are
a~priori unknown. Additionally, the number of solar cells in a \gls{PV} module
and their layout can vary. \change[label=ch:weak,ref=c:weak]{For the estimation
  of lens parameters, the lens distortion of \gls{EL} images of \gls{PV} modules
  may be too weak to apply existing approaches that rely on strong image
deformations to unambiguously deduce the lens parameters.}{We also note that
existing lens models generally assume wide angle lenses. However, their
application to standard lenses is to our knowledge not widely studied.}

To estimate the parameters of a lens distortion model, the plumbline constraint
is typically employed~\cite{Brown1971}. The constraint exploits the fact that
the projection of straight lines under radial and tangential distortion will not
be truly straight. For example, under radial distortion, straight lines are
images as curves. For typical visual inspection tasks, a single image is
sufficient to estimate the lens distortion
parameters~\cite{Devernay2001,Fitzgibbon2001,Claus2005,Ahmed2005,Claus2005a,Rosten2011}.
This can be achieved by decoupling the intrinsic parameters of the camera from
the parameters of the lens distortion model~\cite{Devernay2001}.

Novel methodologies employ \glspl{CNN} for various segmentation tasks. Existing
\gls{CNN}-based segmentation tasks can be categorized into
\begin{enumerate*}[label=(\arabic*)]
  \item object detection,
  \item semantic segmentation, and
  \item instance-aware segmentation.
\end{enumerate*}
One of the first \gls{CNN} object detection architectures is
\gls{R-CNN}~\cite{Girshick2013} to learn features that are subsequently
classified using a class-specific linear \gls{SVM} to generate region proposals.
\Gls{R-CNN} learns to simultaneously classify object proposals and refine their
spatial locations. The predicted regions, however, provide only a coarse
estimation of object's location in terms of bounding boxes. \citet{Girshick2015}
proposed \gls{Fast R-CNN} by accelerating training and testing times while also
increasing the detection accuracy. \citet{Ren2015} introduced \gls{RPN} that
shares full-image convolutional features with the detection network enabling
nearly cost-free region proposals. \Gls{RPN} is combined with \gls{Fast R-CNN}
into a single network that simultaneously predicts object bounds and estimates
the probability of an object for each proposal.
For semantic segmentation, \citet{Long2015} introduced \glspl{FCN} allowing for
pixelwise inference. The \gls{FCN} is learned end-to-end and pixels-to-pixels
requiring appropriately labeled training data.
Particularly, in medical imaging the U-Net network architecture by
\citet{Ronneberger2015} has been successfully
applied for various segmentation tasks.
In instance segmentation, \citet{Li2016} combined segment proposal and object
detection for \gls{FCIS} where the general idea is to predict the locations
in a fully convolutional network.  \citet{He2017} proposed a Mask \gls{R-CNN} which
extends Faster R-CNN.

The work by \citet{Mehta2018} introduces a \gls{CNN} for the prediction of power
loss. Their system additionally localizes and classifies the type of soiling.
Their work is based on RGB images of whole \gls{PV} modules and addresses the
additional geometric challenges of acquisitions in the field.  In contrast, this
work operates on \gls{EL} images of individual cells of a \gls{PV} module, and
in particular focuses on their precise segmentation in a manufacturing setting.

The main limitation of learning-based approaches is the requirement of a
considerable number of appropriately labeled images for training. However,
pixelwise labeling is time-consuming, and in absence of data not possible at
all. Also, such learning-based approaches require training data that is
statistically representative for the test data, which oftentimes requires to
re-train a model on data with different properties. In contrast, the proposed
approach can be readily deployed to robustly segment \gls{EL} images of \gls{PV}
modules without notable requirements of labeled training data.

The closest work related to the proposed method was presented by
\citet{Sovetkin2019}. This method proposes a robust \gls{PV} module grid
alignment for the application on field \gls{EL} images, where radial and
perspective distortion, motion blur, and disturbing background may be present.
The method uses an external checkerboard calibration for radial distortion
correction, and prior knowledge on the solar cell topology in terms of the
relative distances of the grid lines separating the busbars and cell segments.
In contrast, \gls{EL} images taken under manufacturing conditions may be cropped
or rotated, and the camera is not always pre-calibrated. Hence, the proposed
method performs an automated on-line calibration for every \gls{EL} image. This
is particularly useful for \gls{EL} images of \gls{PV} modules from various
sources, for which the camera parameters may not be available, or when zoom
lenses are used. Additionally, the proposed method performs a pixelwise
classification of pixels belonging to the active cell area and therefore is able
to provide masks tailored to a specific module type. Such masks allow to exclude
unwanted background information and to simplify further processing.

\addition[label=a:decoupling-steps,ref=c:fov-end-to-end]{%
In this work, we unify lens distortion estimation and grid detection by building
upon ideas of \citet{Devernay2001}. However, instead of using independent line
segments to estimate lens distortion parameters, we constrain the problem using
domain knowledge by operating on a coherent grid. This joint methodology allows
to correct errors through feedback from the optimization loop used for
estimating lens model parameters. The proposed approach conceptually differs
from \citet{Sovetkin2019}, where both steps are decoupled and an external
calibration is required.}

\section{Methodology}
\label{sec:el-pv-seg-methodology}

The proposed framework uses a bottom-up pipeline to gradually infer a high-level
representation of a solar module and its cells from low-level ridge edge features in
an \gls{EL} image. Cell boundaries and busbars are represented as parabolic
curves to robustly handle radial lens distortion which causes straight lines to
appear curved in the image. Once we estimated the lens distortion parameters,
the parabolas are rectified to obtain a planar cell grid. This rectified
representation is used to segment the solar cells.

\subsection{Overview}

\begin{figure*}[tp]
  \centering
  \tikzsetnextfilename{cell-extraction-pipeline}
  \input{figures/cell-extraction-pipeline}%
  \caption{The proposed \gls{PV} module segmentation pipeline consists of four
    stages. In the preprocessing stage~\subref{fig:el-preprocessing}, local
    ridge features are extracted. In the curve extraction
    stage~\subref{fig:el-curve-extraction}, candidate parabolic curves are
    determined from ridges. In the model estimation
    stage~\subref{fig:el-curve-grid-model}, a coherent grid and the lens
    distortion are jointly estimated. In the cell extraction
    stage~\subref{fig:el-curve-grid-cell-extraction} the cell topology is
  determined and the cells are extracted.}
  \label{fig:el-pv-pipeline}
\end{figure*}

The general framework for segmenting the solar cells in \gls{EL} images of
\gls{PV} modules is illustrated in \cref{fig:el-pv-pipeline} and consists of the
following steps. First, we locate the busbars and the inter solar cell borders
by extracting the ridge edges. The ridge edges are extracted at subpixel
accuracy and approximated by a set of smooth curves defined as second-degree
polynomials. The parametric representation is used to construct an initial grid
of perpendicularly arranged curves that identify the \gls{PV} module. Using this
curve grid, we estimate the initial lens distortion parameters and hypothesize
the optimal set of curves by further excluding outliers in a \gls{RANSAC}
scheme. Then we refine the lens distortion parameters that we eventually use to
rectify the \gls{EL} image. From the final set of curves we infer the \gls{PV}
module configuration and finally extract the size, perspective, and orientation
of solar cells.

\subsection{Preprocessing}
\label{sec:el-pv-seg-preprocessing}

\addition[label=a:underexpose,ref=c:motivation]{First, the contrast of an
  \gls{EL} image is enhanced to account for possible underexposure.} Then,
  low-level edge processing is applied to attenuate structural variations that
  might stem from cracks or silicon wafer texture, with the goal of preserving
  larger lines and curves.

\subsubsection{Contrast Enhancement}

Here, we follow the approach by~\citet{Franken2006}. A copy~\(I_\text{bg}\) of
the input \gls{EL} image~\(I\) is blurred with a Gaussian kernel, and a
morphological closing with a disk-shaped structure element is applied. Dividing
each pixel of~\(I\) by~\(I_\text{bg}\) attenuates unwanted background noise
while emphasizing high contrast regions. Then, histogram
equalization~\cite[pp.~134\,sqq.]{Gonzalez2018} is applied to increase its
overall contrast. \Cref{fig:el-pv-seg-bgeq-example} shows the resulting
image~\(I\).

\subsubsection{Gaussian Scale-Space Ridgeness}
\label{sec:gaussian-scale-space}

\addition[label=a:gauss-ridges,ref=c:motivation]{The high-level grid structure
  of a \gls{PV} module is defined by inter-cell borders and
  busbars, which correspond to ridges in the image. Ridge edges can be determined
  from second-order partial derivatives summarized by a Hessian. To robustly
  extract line and curve ridges, we compute the second-order derivative of the
  image at multiple scales~\cite{Lindeberg1996,Lindeberg1998}. The responses are
  computed in a Gaussian pyramid constructed from an input \gls{EL}
  image~\cite{Lindeberg1994}. This results in several layers of the pyramid at
  varying resolutions commonly referred to as octaves. The eigendecomposition of
  the Hessian computed afterwards provides information about line-like
structures.}

More in detail,
let \( \vec u \coloneqq (u,v)^\transp\) denote discrete pixel coordinates,
\(O\in\mathbb{N}\) the number of octaves in the pyramid, and \(P\in\mathbb{N}\)
the number of sublevels in each octave. At the finest resolution, we
set~\(\sigma\) to the golden ratio \(\sigma=\sfrac{1+\sqrt{5}}{2} \approx 1.6\).
At each octave~\(o\in\{0,\dotsc,O-1\}\) and
sublevel~\(\ell\in\{0,\dotsc,P-1\}\), we compute the Hessian by
convolving the image with the derivatives of the Gaussian kernel. To obtain the
eigenvalues, the symmetric Hessian is diagonalized by annihilating the
off-diagonal elements using the Jacobi method which iteratively applies Givens
rotations to the matrix~\cite{Golub2013}. This way, its eigenvalues and the
corresponding eigenvectors can be simultaneously extracted in a numerically
stable manner. Let \( \tens H = \mat V \mat \Lambda \mat V^{\transp} \) denote
the eigendecomposition of the Hessian~\(\tens H\), where \( \mat
\Lambda \coloneqq \diag(\lambda_1, \lambda_2) \in \mathbb{R}^{2\times 2} \) is a
diagonal matrix of eigenvalues \( \lambda_1 > \lambda_2 \) and \(\tens V
\coloneqq (\vec v_1, \vec v_2)\) are the associated eigenvectors. Under a
Gaussian assumption, the leading eigenvector dominates the likelihood if the
associated leading eigenvalue is spiked. In this sense, the local ridgeness
describes the likelihood of a line segment in the image at position~\( \vec u
\), and the orientation of the associated eigenvector specifies the
complementary angle~\(\beta(\vec u)\) of the most likely line segment
orientation at position~\(\vec u\). The local ridgeness \( R(\vec u)\) is
obtained as the maximum positive eigenvalue~\(\lambda_1(\vec u)\) across all
octaves and sublevels. Both the ridgeness~\(R(\vec u)\) and the
angle~\(\beta(\vec u)\) provide initial cues for ridge edges in the \gls{EL}
image (see \cref{fig:el-pv-seg-ridgeness}).


\subsubsection{Contextual Enhancement via Tensor Voting}
\label{sec:tensor-voting}

Ridgeness can be very noisy (\cf, \cref{fig:el-pv-seg-ridgeness}).  To discern
noise and high curvatures from actual line and curve features, \(R(\vec u)\)~is
contextually enhanced using tensor voting~\cite{Medioni2000}.

Tensor voting uses a stick tensor voting field to model the likelihood that a
feature in the neighborhood belongs to the same curve as the feature in the
origin of the voting field~\cite{Franken2006a}. The parameter
\(\varsigma > 0\) controls the proximity of the voting field, and
\(\nu\)~determines the angular specificity that we set to \( \nu=2 \) in our
experiments.

Following \citet{Franken2006a}, stickness \(\tilde{R}(\vec
u)=\tilde{\lambda}_1-\tilde{\lambda}_2\) is computed as the difference between
the two eigenvalues~\(\tilde{\lambda}_1,\tilde{\lambda}_2\) of the tensor
field, where \(\tilde{\lambda}_1 >\tilde{\lambda}_2\).
\(\tilde{\beta}(\vec u)=\angle\tilde{\vec e}_1\) is the angle of the
eigenvector~\(\tilde{\vec e}_1\in\mathbb{R}^2\) associated with the largest
eigenvalue~\(\tilde{\lambda}_1\), analogously to~\(\beta(\vec u)\).

We iterate tensor voting two times, since one pass is not always
sufficient~\cite{Franken2006}. Unlike~\citeauthor{Franken2006}, however, we do
not thin out the stickness immediately after the first pass to avoid too many
disconnected edges. Given the high resolution of the \gls{EL} images in our
dataset of approximately \(2500\times2000\)~pixels, we use a fairly large
proximity of ~\(\varsigma_1=15\) in the first tensor voting step, and
\(\varsigma_2=10\) in the second.

\Cref{fig:el-pv-seg-stickness} shows a typical stickness~\(\tilde{R}(\vec u)\)
output. The stickness along the orientation~\(\tilde{\beta}(\vec u)\) is used to
extract curves at subpixel accuracy in the next step of the pipeline.

\subsection{Curve Extraction}
\label{sec:el-pv-seg-curve-extraction}

\begin{figure*}[tp]
  \centering
  \tikzsetnextfilename{subpixel-gaussian}
  \includegraphics[width=\linewidth]{figures/subpixel-gaussian}
  \caption{Extraction of ridge edges from stickness at subpixel accuracy.
    \subref{fig:stickness-ridge-centerline} shows a stickness patch with its
    initial centerline~(\ref*{pgfplots:gaussian-centerline}) at discrete
    coordinates obtained by skeletonization. The refined ridge centerline at
    subpixel accuracy is estimated by fitting a Gaussian
    function~(\ref*{pgfplots:profile-gaussian}) to the cross-section profile of
    the ridge edge in~\subref{fig:3d-gaussian-profile} to equidistantly sampled
    stickness values within a predefined sampling
  window~(\ref*{pgfplots:gaussian-window}).}
  \label{fig:subpixel-ridge}
\end{figure*}

\change[label=ch:curve-motivation,ref=c:motivation]
{Centerline points of ridges are grouped by their curvature.} {We seek to obtain
a coherent grid which we define in terms of second-degree curves. These curves
are traced along the previously extracted ridges by grouping centerline points
by their curvature.} We then fit second-degree polynomials to these points,
which yields a compact high-level curve representation while simultaneously
allowing to discard point outliers.

\subsubsection{Extraction of Ridges at Subpixel Accuracy}
\label{sec:ridges-subpixel-accuracy}

To ensure a high estimation accuracy of lens distortion parameters, we extract
ridge edges at subpixel accuracy. This also makes the segmentation more
resilient in out-of-focus scenarios, where images may appear blurry and the
ridge edges more difficult to identify due to their smoother appearance.
\addition[label=a:blurry,ref=c:blurry]{Blurry images can be caused by slight
camera vibrations during the long exposure time of several seconds that is
required for imaging.  Additionally, focusing in a dark room can be challenging,
hence blur cannot be always avoided. Nevertheless, it is beneficial to be able
to operate also on blurry images, as they can still be useful for defect
classification and power yield estimation in cell areas that do not irradiate.}

To this end, we perform non-maximum suppression by Otsu's global
thresholding~\cite{Otsu1975} on the stickness~\(\tilde{R}(\vec u)\) followed by
skeletonization~\cite{Saeed2010}. Afterwards, we collect the points that
represent the centerline of the ridges through edge linking~\cite{Kovesi2017}.
The discrete coordinates can then be refined by setting the centerline to the
mean of a Gaussian function fitted to the edge profile~\cite{Fabijanska2012}
using the \gls{GN} optimization algorithm~\cite{Nocedal2006}. The 1-dimensional
window of the Gaussian is empirically set to 21 pixels, with four sample points
per pixel that are computed via bilinear interpolation. The \gls{GN} algorithm
is initialized with the sample mean and standard deviation in the window, and
multiplicatively scaled to the stickness magnitude at the mean. The mean of the
fitted Gaussian is then reprojected along the edge profile oriented
at~\(\tilde{\beta}(\vec u)\) to obtain the edge subpixel position.
\Cref{fig:subpixel-ridge} visualizes these steps.

\addition[label=a:hyper-free-subpixel,ref={c:hyperparameters}]
{A non-parametric alternative to fitting a Gaussian to the ridge edge profile
constitutes fitting a parabola instead~\cite{Devernay1995}. Such an approach
is very efficient since it involves a closed-form solution. On the downside,
however, the method suffers from oscillatory artifacts which require
additional treatment~\cite{GromponevonGioi2017}.}

\subsubsection{Connecting Larger Curve Segments}
\label{sec:connecting-curve-segments}

A limitation of the edge linking method is that it does not prioritize curve
pairs with similar orientation. To address this, we first reduce the set of
points that constitute a curve to a sparse representation using the
non-parametric variant of the Ramer-Douglas-Peucker
algorithm~\cite{Ramer1972,Douglas1973} introduced by \citet{Prasad2012}.
Afterwards, edges are disconnected if the angle between the corresponding line
segments is nonzero. In a second pass, two line segments are joined if they are
nearby, of approximately the same length, and pointing into the same direction
within an angle range~\(\vartheta=\SI{5}{\degree}\).
\Cref{fig:curves-heuristics} illustrates the way two curve segments are
combined.

In the final step, the resulting \(n_i\)~points of the \(i\)-th curve of a line
segment form a matrix \(\hat{\mat Q}^{(i)} \in \mathbb{R}^{2\times n_i}\). For
brevity, we denote the \(j\)-th column of \(\hat{\mat Q}^{(i)} \) by \(\hat{\vec
q}_j \in \mathbb{R}^2\).  \(\hat{\mat Q}^{(i)}\) is used to find the parametric
curve representation.

\begin{figure}[tp]
  \centering
  \tikzsetnextfilename{curves-heuristics}%
  \input{figures/curves-heuristics}%
  \caption{When considering combining two adjacent curve segments, one with the
    end line segment \( \protect\overrightarrow{AB} \) and the other with the
    start line segment \( \protect\overrightarrow{B'A'} \), we evaluate the
    angles \(\alpha_1\), \(\alpha_2\), and \(\alpha_3\) and ensure they are
  below the predefined threshold~\(\vartheta\) with \(\alpha_1, \alpha_2 \geq
\alpha_3 \geq \pi-\vartheta\). This way, the combined curve segments are ensured
to have a consistent curvature.}
  \label{fig:curves-heuristics}
\end{figure}

\subsubsection{Parametric Curve Representation}
\label{sec:parametric-curves}

Projected lines are represented as second-degree polynomials to model radial
distortion. The curve parameters are computed via linear regression on the curve
points. More specifically, let
\begin{equation}
  f(x) = a_2 x^2+a_1 x + a_0
  \label{eq:horizontal_parabola}
\end{equation}
denote a second-degree polynomial in horizontal or vertical direction.  The
curve is fitted to line segment points \( \hat{\vec q}_j \in \{
	(x_j,y_j)^\transp  \mid  j=1,\dotsc,n_i\} \subseteq \hat{\mat Q}^{(i)} \) of
the \(i\)-th curve \(\hat{\mat Q}^{(i)}\) by minimizing the \gls{MSE}
\begin{equation}
  \text{MSE}(f) = \frac{1}{n_i}  \sum_{j=1}^{n_i} ( f(x_j) - y_j )^2
  \label{eq:horizontal-parabola-mse}
\end{equation}
using \gls{RANSAC} iterations~\cite{Fischler1981}. In one iteration, we randomly
sample three points to fit \cref{eq:horizontal_parabola}, and then determine
which of the remaining points support this curve model via \gls{MSE}. Outlier
points are discarded if the squared difference between the point and the
parabolic curve value at its position exceeds~\(\rho=1.5\). To keep the
computational time low, \gls{RANSAC} is limited to \num{100}~iterations, and
stopped early once sufficiently many inliers at a \SI{99}{\percent} confidence
level are found~\cite[ch.~4.7]{Hartley2004}. After discarding the outliers, each
curve is refitted to supporting candidate points using linear least
squares~\cite{Golub2013}. To ensure a numerically stable and statistically robust
fit, the 2-D coordinates are additionally normalized~\cite{Harker2008}.

\begin{figure*}[tbp]
  \centering
  \tikzsetnextfilename{preprocessing}
  \setkeys{Gin}{width=(\linewidth-1em)/2}%
  \input{figures/preprocessing}
  \caption{Visualization of the preprocessing, curve extraction, and model
  estimation stages for the \gls{PV} module from
\cref{fig:el-pv-image-to-cells}}
  \label{fig:el-pv-seg-steps}
\end{figure*}

\subsection{Curve Grid Model Estimation}
\label{sec:el-pv-seg-lens-distortion-estimation}

The individual curves are used to jointly form a grid, which allows to further
discard outliers, and to estimate lens distortion. To estimate the lens
distortion, we employ the plumbline constraint~\cite{Brown1971}. The constraint
models the assumption that curves in the image correspond to straight lines in
real world. In this way, it becomes possible to estimate distortion efficiently
from a single image, which allows to use this approach also post~hoc on
cropped, zoomed or similarly processed images.

\subsubsection{Representation of Lens Distortion}

Analogously to \citet{Devernay2001}, we represent the radial lens distortion by
a function \( L\colon \mathbb{R}_{\geq 0} \to \mathbb{R}_{\geq 0} \) that maps
the distance of a pixel from the distortion center to a distortion factor. This
factor can be used to radially displace each normalized image coordinate~\(
\tilde{\vec x } \).

Image coordinates are normalized by scaling down coordinates \( \vec
x\coloneqq(x,y)^\transp \) horizontally by the distortion aspect ratio~\(s_x\)
(corresponding to image aspect ratio decoupled from the projection on the image
plane) followed by shifting the center of distortion \( \vec c \coloneqq
(c_x,c_y)^\transp \) to the origin and normalizing the resulting 2-D point to
the unit range using the dimensions~\(M\times N\) of the image of width~\(M\)
and height~\(N\). Homogeneous coordinates allow to express the normalization
conveniently using a matrix product. By defining the upper-triangular matrix
\begin{equation}
  \mat K =
  \begin{bmatrix}
    s_x M & 0 & c_x
    \\
    0 & N & c_y
    \\
    0 & 0 & 1
  \end{bmatrix}
\end{equation}
the normalizing mapping
\( \normcoord\colon \Omega \to [-1,1]^2 \) is
\begin{equation}
  \normcoord(\vec x)
  =
  \proj\left(
  \mat K^{-1}
  \proj^{-1}(\vec x)
  \right)
  \enspace ,
\end{equation}
where \(\proj \colon \mathbb{R}^3 \to \mathbb{R}^2 \) projects
homogeneous to inhomogeneous coordinates,
\begin{equation}
  \proj \colon (x, y, z)^\transp \mapsto \frac{1}{z} (x,y)^\transp
  \enspace,
  \quad
  \text{for}~z\neq0
\end{equation}
and the inverse operation \( \proj^{-1} \colon \mathbb{R}^2 \to \mathbb{R}^3 \)
backprojects inhomogeneous to homogeneous coordinates:
\begin{equation}
  \proj^{-1}\colon (x,y)^\transp \mapsto (x,y,1)^\transp
  \enspace .
\end{equation}
Note that the inverse mapping \(\normcoord^{-1}\) converts normalized image
coordinates to image plane coordinates.

\subsubsection{The Field-of-View Lens Distortion Model}

To describe the radial lens distortion, we use the first-order \gls{FOV} lens
model by \citeauthor{Devernay2001} that has a single distortion
parameter~\(\omega\). While images can also suffer from tangential distortion,
this type of distortion is often negligible~\cite{Tsai1987}. The sole parameter
\(0< \omega \leq \pi\) denotes the opening angle of the lens. The corresponding
radial displacement function~\( L\) is defined in terms of the distortion radius
\( r\geq0 \) as
\begin{equation}
  L(r) = \frac{1}{\omega} \arctan \left( 2 r\tan \frac{\omega}{2} \right)
  \enspace ,
  \quad
  \text{for}~\omega\neq0
  \enspace .
  \label{eq:fov1_radial_displacement}
\end{equation}
One advantage of the model is that its inversion has a closed-form solution with
respect to the distortion radius~\(r\).

Similar to~\citeauthor{Devernay2001}, we decouple the distortion from the
projection onto the image plane, avoiding the need to calibrate for intrinsic
camera parameters. Instead, the distortion parameter~\(\omega\) is combined with
the distortion center~\(\vec c\in\Omega\) and distortion aspect ratio~\(s_x\)
which are collected in a vector~\( \vec \theta \coloneqq (\vec c, s_x, \omega)
\).

Normalized undistorted image coordinates \(\tilde{\vec x}_u = \vec
\delta^{-1}(\tilde{\vec x}_d) \) can be directly computed from distorted
coordinates \( \tilde{\vec x}_d \) as
\begin{equation}
  \undistort(\tilde{\vec x}_d) =
  \frac{L^{-1}(r_d)}{r_d}\tilde{\vec x}_d
  \enspace ,
  \quad
  \text{for}~r_d \neq 0
  \label{eq:fov-undistort}
\end{equation}
where \( r_d = \norm{\tilde{\vec x}_d}_2 \) is the distance
of \( \tilde{\vec x}_d \) from the origin. \( L^{-1}(r) \) is the inverse of the
lens distortion function in~\cref{eq:fov1_radial_displacement}, namely
\begin{equation}
  L^{-1}(r) = \frac{\tan r \omega}{2 \tan \frac{\omega}{2}}
  \enspace ,
  \quad
  \text{for}~\omega\neq0
  \enspace .
  \label{eq:fov1-undistort}
\end{equation}
The function that undistorts a point~\(\vec x\in\Omega\) is thus
\begin{equation}
  \vec u (\vec x)=
  \normcoord^{-1}
  \left(
  \undistort\left(\normcoord (\vec x)\right)\right)
  \enspace .
  \label{eq:el-pv-seg-undistort}
\end{equation}
Note that \cref{eq:fov-undistort} exhibits a singularity at \(r_d\approxeq0\)
for points close to the distortion center. By inspecting the function's limits,
one obtains
\begin{equation}
  \lim_{r_d\to 0^+} \vec
  \delta^{-1}(\tilde{\vec{x}}_d) = \frac{\omega}{2 \tan\frac{\omega}{2}}
  \tilde{\vec{x}}_d
  \enspace.
\end{equation}
Analogously, \cref{eq:fov1-undistort} is singular at \(\omega=0\) but approaches
\(\lim_{r\to 0^+} L^{-1}(r)=r\) at the limit. In this case,
\cref{eq:fov-undistort} is an identity transformation which does not radially
displace points.

\subsubsection{Estimation of Initial Lens Distortion Model Parameters}
\label{sec:initial-params}

Lens distortion is specified by the distortion coefficient~\(\omega\), the
distortion aspect ratio~\( s_x \), and the distortion center~\( \vec c \). Naive
solution leads to a non-convex objective function with several local minima.
Therefore, we first seek an initial set of parameters close to the optimum, and
then proceed using a convex optimization to refine the parameters.

\addition[label=a:init-scheme,ref=c:refine]{We propose the following
initialization scheme for the individual parameters of the \gls{FOV} lens
model.}

\paragraph{Distortion Aspect Ratio and Center}

We initialize the distortion aspect ratio to \(s_x=1\), and the distortion
center to the intersection of two perpendicular curves with smallest
coefficients in the highest order polynomial term. Such curves can be assumed to
have the smallest curvature and are thus located near the distortion center.

To find the intersection of two perpendicular curves, we denote the coefficients
of a horizontal curve by \(a_2,\allowbreak a_1,\allowbreak a_0\), and the
coefficients of a vertical curve by \(b_2,\allowbreak b_1,\allowbreak b_0\). The
position~\(x\) of a curve intersection is then the solution to
\begin{multline}
   a_{2}^{2} b_{2} x^{4} +
   2 a_{1} a_{2} b_{2} x^{3} +
   x^{2} \bigl(2 a_{0} a_{2} b_{2} +
   a_{1}^{2} b_{2} +
   a_{2} b_{1}\bigr) + x\\
   \cdot(2 a_{0} a_{1} b_{2} + a_{1} b_{1} - 1)+
   a_{0}^{2} b_{2} + a_{0} b_{1} +
   b_{0}
   = 0\enspace.
  \label{eq:parabola-intersection}
\end{multline}
The real roots of the quartic~\eqref{eq:parabola-intersection} can be found with
the Jenkins-Traub \textsc{Rpoly} algorithm~\cite{Jenkins1970} or a specialized
quartic solver~\cite{Flocke2015}. The corresponding values~\(f(x)\) are
determined by inserting the roots back into \cref{eq:horizontal_parabola}.

\paragraph{Distortion Coefficient}

Estimation of the distortion coefficient~\(\omega\) from a set of distorted
image points is not straightforward because the distortion function~\(
L(r) \) is non-linear. One way to overcome this problem is to linearize~\(
L(r) \) with Taylor polynomials, and to estimate~\( \omega \) with linear
least squares.

To this end, we define the distortion factor
\begin{equation}
  k \coloneqq \frac{L(r)}{r}
  \enspace,
  \quad
  \text{for}~k\in\mathbb{R}_{>0}
  \label{eq:dist-factor}
\end{equation}
which maps undistorted image points \( \{\vec p_j\}_{j=1}^n \) lying on the
straight lines to distorted image points \( \{\vec q_j\}_{j=1}^n
\) lying on the parabolic curves. Both point sets are then related by
\begin{equation}
  \vec p k = \vec q
  \enspace
  .
\end{equation}
The distorted points \( \vec q_j \) are straightforward to extract by evaluating
the second-degree polynomial of the parabolic curves. To determine~\( \vec p_j
\), we define a line with the first and the last point in~\(\vec q_j \), and
select points from this line. Collecting these points in the vectors \( \vec p
\in \mathbb{R}^{2n} \) and \( \vec q \in \mathbb{R}^{2n} \) yields an
overdetermined system of \(2n\)~linear equations in one unknown. \(\hat k\)~is
then estimated via linear least squares as
\begin{equation}
  \hat k = \argmin_{k} 
  \norm{\vec q - \vec p k}_2^2\enspace,
\end{equation}
where the solution is found via the normal equations~\cite{Golub2013} as
\begin{equation}
  \hat k \coloneqq \frac{\vec p^\transp \vec q}{\vec p^\transp \vec p}
  \enspace.
\end{equation}
The points \(\vec q_j,\vec p_j\) refer to the columns of the two matrices \(\mat
Q^{(i)}, \mat P^{(i)} \in \mathbb{R}^{2\times n_i}\), respectively, where
\(n_i\) again denotes the number of points, which are used in the following step
of the pipeline.

To determine~\( \omega \) from the relation \( k=\frac{L(r)}{r} \), \( L(r) \)
is expanded around~\(\omega_0=0\) using Taylor series. More specifically, we use
a second-order Taylor expansion to approximate
\begin{equation}
\arctan(x) =  x+\mathcal{O}(x^2)\enspace,
\label{eq:arctan-approx}
\end{equation}
and a sixth-order Taylor expansion to
approximate
\begin{equation}
\tan(y) = y+\frac{y^3}{3}+\frac{2y^5}{15} + \mathcal{O}(y^6) \enspace.
\label{eq:tan-approx}
\end{equation}
Let \( L(r)= \frac{1}{\omega}\arctan(x) \) with \(x=2r \tan (y)\), and \(
y=\frac{\omega}{2}\). We substitute the Taylor polynomials from
\cref{eq:arctan-approx,eq:tan-approx}, and \(x,y\) into \cref{eq:dist-factor} to
obtain a biquadratic polynomial~\( Q(\omega)\) independent of~\( r \):
\begin{equation}
  \frac{L(r)}{r}
  \approx \underbrace{1 + \frac{1}{12}\omega^2 + \frac{1}{120}
\omega^4}_{\eqqcolon Q(\omega)}\enspace.
  \label{eq:L_approx}
\end{equation}
By equating the right-hand side of \cref{eq:L_approx} to~\( k\)
\begin{equation}
  Q(\omega) = k
  \label{eq:distort_approx_poly}
\end{equation}
we can estimate~\( \omega \) from four roots of the resulting
polynomial~\(Q(\omega)\). These roots can be found by substituting~\( z=\omega^2
\) into~\cref{eq:L_approx}, solving the quadratic equation with respect to~\(
z\), and substituting back to obtain~\( \omega \). This eventually results in
the four solutions~\(\pm \sqrt{z_{1,2}} \). The solution exists only if~\( k
\geq 1 \), as complex solutions are not meaningful, and thus corresponds to the
largest positive real root.

\begin{figure}[tb]
  \centering
  \tikzsetnextfilename{omega-approximation}
  \includegraphics[width=\linewidth]{figures/omega-approximation}%
  \caption{Approximation of the distortion coefficient~\( \omega \) using
    \cref{eq:L_approx}~(\ref*{pgfplots:omega-approx}) compared to the exact
    solution with respect to varying radii~\( r \). For large radii outside the
    range of normalized coordinates (\ie, the radius of the half-unit circle~\(
    r > \sfrac{1}{\sqrt{2}} \)), the estimate is not accurate. This implies that
    the ideal sampled points must be both at some distance from the image border
    and also from the distortion center. As a side note, the estimation error
  becomes unacceptable for wide lenses where~\(\omega > \sfrac{\pi}{4} \).
However, the \gls{EL} images in this work~(\(\bar{\omega}\)) are well below this
threshold.}
  \label{fig:el-pv-seg-omega-estimate}
\end{figure}

We evaluated the accuracy of the approximation~\eqref{eq:L_approx} with the
results shown in \cref{fig:el-pv-seg-omega-estimate}. For large radii, the
approximation significantly deviates from the exact solution. Consequently, this
means that the selected points for the estimation must ideally be well
distributed across the image. Otherwise, the lens distortion parameter will be
underestimated. In practice, however, this constraint does not pose an issue due
to the spatial distribution of the solar cells across the captured \gls{EL}
image.

\subsubsection{Minimization Criterion for the Refinement of Lens Distortion Parameters}
\label{sec:refinement}

The Levenberg-Marquardt algorithm~\cite{Levenberg1944, Marquardt1963} is used to
refine the estimated lens distortion parameters~\(\vec\theta\). The objective
function is
\begin{equation}
  \vec\theta^\star \coloneqq
  \argmin_{\vec\theta}\frac{1}{2}\sum_{i=1}^n \chi^2(\mat P^{(i)},\vec\theta)
  \enspace .
  \label{eq:cost-function}
\end{equation}
\(\mat P^{(i)}
\in\mathbb{R}^{2\times m} \) is a matrix of \(m\) 2-D points of the \(i\)-th
curve. The distortion error~\(\chi^2\) quantifies the deviation of the points
from the corresponding ideal straight line~\cite{Devernay2001}. The undistorted
image coordinates \(\vec p_j\coloneqq(x_j,y_j)^\transp\in\Omega\) are computed
as \( \vec p_j=\vec u(\vec q_j) \) by applying the inverse lens distortion given
in \cref{eq:el-pv-seg-undistort} to the points \(\vec q_j\) of the \(i\)-the
curve \(\mat Q^{(i)} \). In a similar manner, the obtained points \(\vec p_j \)
form the columns of \( \mat P^{(i)} \in \mathbb{R}^{2\times n_i}\).

Following~\citeauthor{Devernay2001}, we iteratively optimize the set of lens
parameters~\(\vec\theta\).  In every step~\(t\), we refine these parameters and
then compute the overall error \(\epsilon_{t}\coloneqq\sum_{i=1}^n\chi^2(\mat
P^{(i)}, \vec\theta)\) over all curve
points. Afterwards, we undistort the curve points and continue the optimization
until the relative change in error~\(\epsilon\coloneqq
(\epsilon_{t-1}-\epsilon_{t}) / \epsilon_t\) falls below the threshold
\(\epsilon=10^{-6}\).

Minimizing the objective function~\eqref{eq:cost-function} for all parameters
simultaneously may cause the optimizer to be trapped in a local minimum.
\change[label=ch:param-subsets,ref=c:refine]{We therefore optimize a subset of
  parameters in several partitions} {Hence, following \citet{Devernay2001}, we
optimize the parameters~\(\vec\theta=(\omega,s_x,\vec c)\) in subsets} starting
with~\(\omega\) only. Afterwards, we additionally optimize the distortion
center~\(\vec c\). Finally, the parameters~\(\vec\theta\) are jointly optimized.

\subsubsection{Obtaining a Consistent Parabolic Curve Grid Model}

The layout of the curves is constrained to a grid in order to eliminate outlier
curves. Ideally, each horizontally oriented parabola should intersect each
vertically oriented parabola exactly once. This intersection can be found using
\cref{eq:parabola-intersection}. Also, every parabolic curve should not
intersect other parabolic curves of same orientation within the image plane.
This set of rules eliminates most of the outliers.

\begin{figure}[tb]
  \centering
  \tikzsetnextfilename{cell-segments}
  \setkeys{Gin}{width=(\linewidth-1em*2)/3}%
  \input{figures/cell-segments}%
  \caption{Estimation of solar module topology requires determining the number
    of subdivisions (\ie, rectangular segments) in a solar cell. Common
    configurations include no subdivisions at all (\ie, one
    segment)~\subref{fig:no-subdivisions}, three
  segments~\subref{fig:three-segments} and four
segments~\subref{fig:four-segments}. Notice how the arrangement of rectangular
segments is symmetric and segment sizes increase monotonically towards the
center, \ie, \( \Delta_1 < \dotsb < \Delta_n\). In particular, shape symmetry
can be observed not only along the vertical axis of the solar cell but also
along the horizontal one as well.}
  \label{fig:cell-segments}
\end{figure}

\begin{figure*}[tb]
  \centering
  \tikzsetnextfilename{mask-generation}%
  \input{figures/mask-generation}%
  \caption{Intermediate steps of the solar mask estimation process}
  \label{fig:el-pv-seg-solar-cell-mask}
\end{figure*}

\paragraph{Robust Outlier Elimination}

\gls{LO-RANSAC}~\cite{Chum2003} is used to remove outlier curves.
In every  \gls{LO-RANSAC} iteration, the grid constraints are imposed by
randomly selecting two horizontal and two vertical curves to build a minimal
grid model. Inliers are all curves that
\begin{enumerate*}[label=(\arabic*)]
  \item exactly once intersect the model grid lines of perpendicular
    orientation,
  \item not intersect the model grid lines of parallel orientation, and
  \item whose \gls{MSE} of the reprojected undistorted points is not larger than
    one pixel.
\end{enumerate*}

\paragraph{Remaining Curve Outliers}

Halos around the solar modules and holding mounts (such as in
\cref{fig:el-pv-seg-steps}) can generate additional curves outside of the cells.
We apply Otsu's thresholding~\cite{Otsu1975} on the contrast-normalized image
and discard outer curves that generate additional grid rows or columns with an
average intensity in the enclosed region below the automatically determined
threshold.

\subsection{Estimation of the Solar Module Topology}
\label{sec:el-pv-seg-solar-module-configuration}

A topology constraint on the solar cell can be employed to eliminate remaining
non-cell curves in the background of the \gls{PV} module, and the number and
layout of solar cells can be subsequently estimated. However, outliers prevent a
direct estimation of the number of solar cell rows and columns in a \gls{PV}
module. Additionally, the number and orientation of segments dividing each solar
cell are generally unknown. Given the aspect ratio of solar cells in the imaged
\gls{PV} module, the topology can be inferred from the distribution of parabolic
curves. For instance, in \gls{PV} modules with equally long horizontal and
vertical cell boundary lines, the solar cells have a square (\ie, \(1:1\))
aspect ratio.

The number of curves crossing each square image area of solar cell is constant.
Clustering the distances between the curves allows to deduce the number of
subdivisions within solar cells.

\subsubsection{Estimation of the Solar Cell Subdivisions and the Number of Rows and
Columns}
\label{sec:solar-cell-subdivisions}

The solar cells and their layout are inferred from the statistics of the line
segment lengths in horizontal and vertical direction. We collect these lengths
separately for each dimension and cluster them. \textsc{Dbscan}
clustering~\cite{Ester1996} is used to simultaneously estimate cluster
membership and the number of clusters. Despite the presence of outlier curves,
clusters are representative of the distribution of segment dimensions within a
cell. For example, if a solar cell consists of three vertically arranged
segments (as in \cref{fig:three-segments}) with heights of \(20:60:20\) pixels,
the two largest clusters will have the medians 60 and 20. With the assumption
that the segment arrangement is typically symmetric, the number of segments is
estimated as the number of clusters times two minus one.  If clustering yields a
single cluster, we assume that the solar cells consist of a single segment.
Outlier curves or segments, respectively, are rejected by only considering the
largest clusters, with the additional constraint that the sizes of the used
clusters are proportional to each other, and that not more than two different
segments (as in \cref{fig:four-segments}) can be expected in a cell. The number
of rows and columns of a solar cell is determined by dividing the overall size
of the curve grid by the estimated cell side lengths.

\subsubsection{Curve Grid Outlier Elimination}

The estimated proportions are used to generate a synthetic planar grid that is
registered against the curve grid intersections. Specifically, we use the rigid
point set registration of \gls{CPD}~\cite{Myronenko2010}
\change[label=ch:cpd,ref=c:motivation]{The method requires choosing a
weight~\(0\leq w \leq 1\) of the uniform distribution that accounts for noise
and outliers, which we estimate as the proportion}{because it is deterministic
and allows us to account for the proportion of outliers using a
parameter~\(0\leq w \leq 1\). We can immediately estimate~\(w\) as the fraction}
of points in the synthetic planar grid and the total number of intersections in
the curve grid.

To ensure \gls{CPD} convergence, initial positions of the synthetic planar grid
should be sufficiently close to the curve grid intersections. We therefore
estimate the translation and rotation of the planar grid to closely pre-align it
with the grid we are registering against. The initial translation can be
estimated as the curve grid intersection point closest to the image plane
origin. The 2-D in-plane rotation is estimated from the average differences of
two consecutive intersection points along each curve grid row and column. This
results in two 2-D vectors which are approximately orthogonal to each other. The
2-D vector with the larger absolute angle is rotated by \SI{90}{\degree} such
that both vectors become roughly parallel. The estimated rotation is finally
obtained as the average angle of both vectors.

\subsubsection{Undistortion and Rectification}
\label{sec:undistort-rectify}

The \gls{PV} module configuration is used to undistort the whole image using
\cref{eq:el-pv-seg-undistort}. After eliminating the lens distortion, we use
\gls{DLT}~\cite{Hartley2004} to estimate the planar 2-D homography using the
four corners of the curve grid with respect to the corners of the synthetic
planar grid. \addition[label=a:homography,ref=c:motivation]{The homography is
used to remove perspective distortion from the undistorted curve grid.}

The intersections of the perspective corrected curve grid may not align exactly
with respect to the synthetic planar grid
\addition[label=a:mls,ref=c:motivation]{because individual solar cells are not
always accurately placed in a perfect grid but rather with a margin of error}.
The remaining misalignment is therefore corrected via affine
\gls{MLS}~\cite{Schaefer2006}, which warps the image using the planar grid
intersections as control points distorted using the estimated lens parameters,
and curve grid intersections are used as their target positions.

\subsection{Estimation of the Active Solar Cell Area}
\label{sec:el-pv-seg-cell-mask}

We use solar cell images extracted from individual \gls{PV} modules to generate
a mask that represents the active solar cell area. Such masks allow to exclude
the background and the busbars of a solar cell (see
\cref{fig:el-pv-seg-solar-cell-mask}). In particular, active cell area masks are
useful for detection of cell cracks since they allow to mask out the busbars,
which can be incorrectly identified as cell cracks due to high similarity of
their appearance~\cite{Spataru2016,Stromer2019}.

Estimation of solar cell masks is related to the image labeling problem, where
the goal is to classify every pixel into several predefined classes (in our
case, the background and the active cell area). Existing approaches solve this
problem using probabilistic graphical models, such as a \gls{CRF} which learns
the mapping in a supervised manner through contextual information~\cite{he2004}.
However, since the estimated curve grid already provides a global context, we
tackle the pixelwise classification as a combination of adaptive thresholding
and prior knowledge with regard to the straight shape of solar cells. Compared
to \glspl{CRF}, this approach does not require a training step and is easy to
implement.

To this end, we use solar cells extracted from a \gls{PV} module to compute a
mean solar cell (see
\crefrange{fig:el-pv-seg-segmented-cells}{fig:el-pv-seg-average-cell}).
\addition[label=a:threshold,ref=c:motivation]{Since intensities within a mean
solar cell image can exhibit a large range}, we apply locally adaptive
thresholding~\cite{Pitas1993} on \(25\times 25\)~pixels patches using their mean
intensity, followed by a \(15\times15\) morphological opening and flood filling
to close any remaining holes. This leads to an initial binary mask.

Ragged edges at the contour are removed using vertical and horizontal cell
profiles (\cref{fig:el-pv-seg-average-cell}). The profiles are computed as
pixelwise median of the initial mask along each image row or column,
respectively. We combine the backprojection of these profiles with the convex
hull of the binary mask determined with the method of~\citet{Barber1996} to
account for cut-off corners using \bitwise{and}~(\cf,
\cref{fig:el-pv-seg-augmented-mask}). To further exclude repetitive patterns in
the \gls{EL} image of a solar cell, \eg., due to low passivation efficiency in
the contact region (see \cref{fig:el-pv-seg-extra-geom}), we combine the initial
binary mask and the augmented mask via \bitwise{xor}.


We note that solar cells are usually symmetric about both axes. Thus, the active
solar cell area mask estimation can be restricted to only on quadrant of the
average solar cell image to enforce mask symmetry. Additionally, the convex hull
of the solar cell and its extra geometry can approximated by
polygons~\cite{Aggarwal1985} for a more compact representation.

\subsection{Parameter Tuning}
\label{sec:parameter-selection}

\begin{table*}[bp]
  \centering
  \caption{Overview of segmentation pipeline parameters and their values used in
  this work}
  \label{tab:pipeline-parameters}
  \begin{tabularx}{\textwidth}{lrXl}
    \toprule
    \multicolumn{1}{c}{\S}
    & Symbol & Description & Used value
    \\
    \midrule
    \multirow{4}{*}{\ref{sec:gaussian-scale-space}}
    &
    \(O\) & Number of octaves in Gaussian scale-space pyramid & 5
    \\
    &
    \(P\) & Number of sublevels in each octave & 8
    \\
    &
    \(\sigma\) & Gaussian scale-space standard deviation & 1.6
    \\
    &
    \(\gamma\) & Gaussian scale-space pyramid downsampling factor & 2
    \\
    \midrule
    \multirow{2}{*}{\ref{sec:tensor-voting}}
    &
    \(\nu\) & Tensor voting angular specificity & 2
    \\
    &
    \(\varsigma_{1,2}\) & Proximity of the 1\textsuperscript{st} and
    2\textsuperscript{nd} tensor voting steps & \(15,10\)
    \\
    \midrule
    \ref{sec:ridges-subpixel-accuracy}
    &
    & 1-D sampling window for Gaussian-based subpixel position & 21
    \\
    \ref{sec:connecting-curve-segments}
    &
    \(\vartheta\)
    &
    Maximum merge angle of two neighboring line segments & \SI{5}{\degree}
    \\
    \midrule
    \ref{sec:parametric-curves}
    &
    \(\rho\) & Maximum error between fitted parabolic curve value at curve point &
    \(1.5\)
    \\
    \ref{sec:refinement}
    &
    \(\epsilon\) & Minimal change in error during refinement of lens distortion
    parameters& \(10^{-6}\)
    \\
    \midrule
    \ref{sec:el-pv-seg-solar-module-configuration}
    &
    & Solar cell aspect ratio & \(1:1\)
    \\
    \ref{sec:el-pv-seg-cell-mask}
    &
    & Locally adaptive thresholding window size & \(25\times25\)
    \\
    \bottomrule
  \end{tabularx}
\end{table*}

The proposed solar cell segmentation pipeline relies on a set of hyperparameters
that directly affect the segmentation robustness and accuracy.
\Cref{tab:pipeline-parameters} provides an overview of all parameters with their
values used in this work.

\subsubsection{Manual Search}
\label{sec:hyper-manual-search}

Since the parameters of the proposed segmentation are intuitive and easily
interpretable, it is straightforward to select them based on the setup used for
\gls{EL} image acquisition.

Main influence factors that must be considered when choosing the parameters are
image resolution and physical properties of the camera lens.

Provided parameter values were found to work particularly well for high
resolution \gls{EL} images and standard camera lenses, as in our dataset (\cf,
\cref{sec:el-pv-dataset}). For low resolution \gls{EL} images, however, the
number of pyramid octaves and sublevels will need to be increased to avoid
missing important image details. Whereas, tensor voting proximity, on contrary,
will need to be lowered, since the width of ridge edges in low resolution images
tends to be proportional to the image resolution. This immediately affects the
size of the 1-D sampling window for determining the Gaussian-based subpixel
position of curve points.

Curve extraction parameters correlate with the field-of-view of the \gls{EL}
camera lens. In particular for wide angle lenses, the merge angle~\(\vartheta\)
must be increased.

Parabolic curve fit error~\(\rho\) balances between robustness and accuracy of
the segmentation result. The window size for locally adaptive thresholding used
for estimation of solar cell masks correlates both with the resolution of
\gls{EL} images, but also with the amount of noise and texture variety in solar
cells, \eg due to cell cracks.

\begin{additionenv}[label=a:auto-tune,ref={c:hyperparameters}]

\subsubsection{Automatic Search}
\label{sec:hyper-auto-search}

The parameters can also be automatically optimized in an efficient manner using
random search~\cite{Rastrigin1963,Schumer1968,Schrack1976,Masri1980,
Solis1981,Bergstra2012} or Bayesian
optimization~\cite{Kushner1964,Mockus1975,Bergstra2011,Bergstra2013,Snoek2012,Akiba2019}
class of algorithms. Since this step involves supervision, pixelwise \gls{PV}
module annotations are needed. In certain cases, however, it may be not be
possible to provide such annotations because individual defective \gls{PV} cells
can be hard to delineate, \eg, they appear completely dark. Also, the active
solar cell area of defective cells is not always well-defined. Therefore, we
refrained from automatically optimizing the hyperparameters in this work.

\end{additionenv}

\section{Evaluation}
\label{sec:el-pv-seg-evaluation}

We evaluate the robustness and accuracy of our approach against manually
annotated ground truth masks. Further, we compare the proposed approach against
the method by \citet{Sovetkin2019} on simplified masks, provide qualitative
results and runtimes, and discuss limitations.

\subsection{Dataset}
\label{sec:el-pv-dataset}

We use a dataset consisting of \num{44}~unique \gls{PV} modules with various
degrees of defects to manually select the parameters for the segmentation
pipeline \addition[label=a:training,ref=c:training]{and validate the results.
  These images served as a reference during the development of the proposed
method}. \addition[label=a:dataset-conditions,ref=c:contrib2]{The \gls{PV}
modules were captured in a testing laboratory setting at different orientations
and using varying camera settings, such as exposure time. Some of \gls{EL}
images were post-processed by cropping, scaling, or rotation.} This
dataset consists of \num{26}~monocrystalline and \num{18}~polycrystalline
solar cells.
In total, these \num{44}~solar modules consist of \num{2624}
solar cells out of which \num{715} are definitely defective with defects ranging
from microcracks to completely disconnected cells and mechanically induced
cracks (\eg, electrically insulated or conducting cracks,
or cell cracks due to soldering~\cite{Spertino2015}). \num{106}~solar cells
exhibit smaller defects that are not with certainty identifiable as completely
defective, and \num{295}~solar cells feature miscellaneous surface abnormalities
that are no defects. The remaining \num{1508}~solar cells are categorized as
functional without any perceivable surface abnormalities. The solar cells in
imaged \gls{PV} modules have a square aspect ratio (\ie, are quadratic).

The average resolution of the \gls{EL} images is
\(\num{2779.630435}\times\num{2087.347826}\) pixels with a standard deviation
of image width and height of \num{576.420115} and \num{198.298565} pixels,
respectively. The median resolution is~\(\num{3152.0}\times \num{2046.0} \)
pixels.

Additional eight test \gls{EL} images
\addition[label=a:test-set-size,ref=c:training]{(\ie,
about \(\SI{15}{\percent}\) of the dataset)} are used for the
evaluation. Four modules are monocrystalline and the remaining four are
polycrystalline. Their ground truth segmentation masks consist of hand-labeled
solar cell segments. The ground truth additionally specifies both the rows and
columns of the solar cells, and their subdivisions. These images show various
\gls{PV} modules with a total of \num{408}~solar cells.  The resolution of the
test \gls{EL} images varies around \(\num{2649.5}\pm \num{643.197592} \times
\num{2074.0}\pm \num{339.124924} \) with a median image resolution
of~\(\num{2581.5}\times \num{2046.0} \).

Three out of four monocrystalline modules consist of \(4\times 9\)~cells and the
remaining monocrystalline module consists of \(6\times10\)~cells.  All of their
cells are subdivided by busbars into \(3\times1\)~segments.

The polycrystalline modules consist of \(6\times 10\)~solar cells each.  In two
of the modules, every cell is subdivided into \(3\times 1\)~segments. The cells
of the other two modules are subdivided into \( 4\times 1\)~segments.

\subsection{Evaluation Metrics}
\label{sec:metrics}

We use two different metrics, pixelwise scores and the weighted Jaccard index to
evaluate both the robustness and the accuracy of the proposed method and to
compare our method against related work. In the latter case, we additionally use
a third metric, the \gls{RMSE}, to compute the segmentation error on simplified
masks.

\subsubsection{Root Mean Square Error}

The first performance metric is the \gls{RMSE} given in pixels between the
corners of the quadrilateral mask computed from the ground truth annotations and
the corners estimated by the individual modalities. The metric provides a
summary of the method's accuracy in absolute terms across all experiments.

\subsubsection{Pixelwise Classification}

The second set of performance metrics are precision, recall, and the \(F_1\)
score~\cite{Rijsbergen1979}. These metrics are computed by considering cell
segmentation as a \change[label=ch:multiclass,ref=c:pixelwise]{binary pixelwise
  classification into active cell area and background pixels}{multiclass
pixelwise classification into background and active area of individual solar
cells. A typical 60~cell \gls{PV} module will therefore contain up to 61 class
labels.} A correctly segmented active area pixel is a true positive, the
remaining quantities are defined accordingly.
\deletion[label=d:binary,ref=c:pixelwise]{It is worth mentioning that these
  metrics do not explicitly penalize mismatches between major segmentation
  errors. For example, if a segmentation merges two separate cells into one,
  pixels from both solar cells are counted as true positives. Yet only pixels
  corresponding to the background between these cells are considered as false
  negatives instead of the whole segment that was merged incorrectly.}
  \addition[label=a:pixelwise-proportion,ref=c:pixelwise]{Pixelwise scores are
    computed globally with respect to all the pixels. Therefore, the differences
  between the individual results for these scores are naturally smaller than for
metrics that are computed with respect to individual solar cells, such as the
Jaccard index.}

\subsubsection{Weighted Jaccard Index}

The third performance metric is the weighted Jaccard
index~\cite{Chierichetti2010,Ioffe2010}, a variant of the metric widely known
as \gls{IoU}. This metric extends the common Jaccard index by an importance
weighting of the input pixels. As the compared masks are not strictly binary
either due to antialiasing or interpolation during mask construction, we define
importance of pixels by their intensity. Given two non-binary masks~\(A\)
and~\(B\), the weighted Jaccard similarity is
\begin{equation}
  J_w
  =
  \frac{\sum_{\vec u \in\Omega} \min \{ A(\vec u), B(\vec u) \} }
  {\sum_{\vec u \in\Omega} \max \{ A(\vec u), B(\vec u) \} }
  \enspace
  .
\end{equation}
The performance metric is computed on pairs of segmented cells and ground truth
masks.  A ground truth cell mask is matched to the segmented cell with the
largest intersection area, thus taking structural coherence into account.

We additionally compute the Jaccard index of the background, which corresponds to
the accuracy of the method to segment the whole solar module. Solar cell
misalignment or missed cells will therefore penalize the segmentation accuracy
to a high degree. Therefore, the solar module Jaccard index provides a summary
of how well the segmentation performs per \gls{EL} image.

\subsection{Quantitative Results}
\label{sec:quantitative}

We evaluate \change[label=ch:robustness,ref=c:contrib2]{the segmentation
accuracy}{the segmentation accuracy and the robustness} of our approach
using a fixed set of parameters as specified in \cref{tab:pipeline-parameters}
\addition[label=a:eval-setting,ref=c:contrib2]{on \gls{EL} images of \gls{PV}
modules acquired in a material testing laboratory}.

\subsubsection{Comparison to Related Work with Simplified Cell Masks}

The method by \citeauthor{Sovetkin2019} focuses on the estimation of the
perspective transformation of the solar module and the extraction of solar
cells. Radial distortion is corrected with a lens model of an external
checkerboard calibration. The grid structure is fitted using a~priori knowledge
of the module topology. For this reason, we refer to the method as \gls{PGA}.
The method makes no specific proposal for mask generation and therefore yields
rectangular solar cells.


\begin{figure*}[tp]
  \centering
  \tikzsetnextfilename{exact-vs-quad-masks}
  \pgfmathsetlengthmacro\imagewidth{(\linewidth-1em)/2}
  \setkeys{Gin}{height=\imagewidth}%
  \input{figures/exact-vs-quad-masks}%
  \caption{Example of an exact mask~\subref{fig:em} of solar cells estimated
    using the proposed approach and a quadrilateral mask~\subref{fig:qm}
    determined from the exact mask. The latter is used for comparison against
    the method of \citet{Sovetkin2019}. Both masks are shown as color overlays.
  Different colors denote different instances of solar cells.}
  \label{fig:em-vs-qm}
\end{figure*}

In order to perform a comparison, the exact masks (\cf, \cref{fig:em}) are
restricted to quadrilateral shapes (\cf, \cref{fig:qm}).
The quadrilateral mask is computed as the minimum
circumscribing polygon with four sides, \ie, a quadrilateral, using the approach
of~\citet{Aggarwal1985}. The quadrilateral exactly circumscribes the convex hull
of the solar cell mask with all the quadrilateral sides flush to the convex
hull.

\gls{PGA} assumes that radial distortion is corrected by an external
checkerboard calibration. This can be a limiting factor in practice. Hence, the
comparison below considers both practical situations by running \gls{PGA} on
distorted images and on undistorted images using the distortion correction of
this work.

\paragraph{Root Mean Square Error}

\pgfplotstableset{
    highlight cell/.style={
      bold cell content,
    },
}

\begin{table*}[tb]
  \centering
  \caption{\Glsfirst{RMSE}, in pixels, of the distance between the corners of
  the quadrilateral mask determined from the ground truth annotations and the
corners determined by the respective method in all eight test images. Bold face
denotes smallest error.}
  \label{tab:rmse}
  \pgfplotstableset{search path={./data},col sep=space}
  \pgfplotstableread{rmse.csv}\rmse
  \pgfplotstabletypeset
  [
    begin table=\begin{tabularx}{\linewidth},
    end table=\end{tabularx},
    every head row/.style={ablation header},
    every last row/.style={
      before row=\midrule,
      after row=\bottomrule,
    },
    multirow header/.style={
      assign cell content/.code={%
        \ifnum\pgfplotstablerow=0
          \pgfkeyssetvalue{/pgfplots/table/@cell content}%
            {\multirow{#1}{*}{a##1}}%
        \else
          \pgfkeyssetvalue{/pgfplots/table/@cell content}{}%
        \fi
      },
    },
    precision=2,
    fixed,
    fixed zerofill,
    columns={
      type,
      pcga-qm-d,
      cgs-qm-d-w-only-no-mls,
      cgs-qm-d-no-mls,
      cgs-qm-d-full,
      pcga-qm-u,
      cgs-qm-u-w-only-no-mls,
      cgs-qm-u-no-mls,
      cgs-qm-u-full
    },
    columns/type/.style={
      string replace={mono}{Monocrystalline},
      string replace={poly}{Polycrystalline},
      string replace={overall}{Overall},
      column type=r,
      column name=,
      string type,
    },
    columns/pcga-qm-d/.style={
      column name=\empty,
      column type=Y,
    },
    columns/cgs-qm-d-w-only-no-mls/.style={
      column name=\(\omega\),
      column type=Y,
    },
    columns/cgs-qm-d-no-mls/.style={
      column name={\((\omega,s_x,\vec c)\)},
      column type=Y,
    },
    columns/cgs-qm-d-full/.style={
      column name={w/~MLS},
      column type=Y,
    },
    every row 0 column 3/.style={
      highlight cell,
    },
    every row 1 column 4/.style={
      highlight cell,
    },
    every row 2 column 4/.style={
      highlight cell,
    },
    columns/pcga-qm-u/.style={
      column name=\empty,
      column type=Y,
    },
    columns/cgs-qm-u-w-only-no-mls/.style={
      column name=\(\omega\),
      column type=Y,
    },
    columns/cgs-qm-u-no-mls/.style={
      column name={\((\omega,s_x,\vec c)\)},
      column type=Y,
    },
    columns/cgs-qm-u-full/.style={
      column name={w/~MLS},
      column type=Y,
      bold cell content,
    },
  ]
  \rmse
\end{table*}

\Cref{tab:rmse} provides the \gls{RMSE} in pixels between the corners of the
quadrilaterals computed by the respective modality and the quadrilateral mask
estimated from the ground truth.  The metric is provided for monocrystalline and
polycrystalline solar wafers separately, and for both types combined. In all
cases, the proposed approach outperforms both \gls{PGA} variants. We
particularly notice that \Gls{PGA} greatly benefits from lens distortion
estimation. This underlines our observation that the latter is essential for
highly accurate segmentation.

\paragraph{Pixelwise Classification}

\begin{table*}[tbp]
  \centering
  \caption{Pixelwise classification scores for quadrilateral masks estimated
  using \gls{PGA} and the proposed approach. Bold face denotes the best
  performing method.}
  \label{tab:pixelwise-qm}
  \pgfplotstableset{
    col sep=comma,
    search path={data/pixelwise_scores},
  }%
  \pgfplotstableread{cgs_qm_d_full.csv}\cgsqmdfull
  \pgfplotstableread{cgs_qm_d_full_no_mls.csv}\cgsqmdfullnomls
  \pgfplotstableread{cgs_qm_d_w_only_no_mls.csv}\cgsqmdwonly
  \pgfplotstableread{cgs_qm_u_full.csv}\cgsqmufull
  \pgfplotstableread{cgs_qm_u_full_no_mls.csv}\cgsqmufullnomls
  \pgfplotstableread{cgs_qm_u_w_only_no_mls.csv}\cgsqmuwonly
  \pgfplotstableread{pcga_qm_u.csv}\pcgaqmu
  \pgfplotstableread{pcga_qm_d.csv}\pcgaqmd
  \pgfplotstabletranspose[transposed scores]\cgsqmdfullnomlst\cgsqmdfullnomls
  \pgfplotstabletranspose[transposed scores]\cgsqmdt\cgsqmdfull
  \pgfplotstabletranspose[transposed scores]\cgsqmdwonlyt\cgsqmdwonly

  \pgfplotstabletranspose[transposed scores]\cgsqmufullnomlst\cgsqmufullnomls
  \pgfplotstabletranspose[transposed scores]\cgsqmut\cgsqmufull
  \pgfplotstabletranspose[transposed scores]\cgsqmuwonlyt\cgsqmuwonly
  \pgfplotstabletranspose[transposed scores]\pcgaqmut\pcgaqmu
  \pgfplotstabletranspose[transposed scores]\pcgaqmdt\pcgaqmd
  \pgfplotstableset{
    combine tables column/.style={
      create on use/pcga-d/.style={
        create col/copy column from table={\pcgaqmdt}{#1},
      },
      create on use/pcga-u/.style={
        create col/copy column from table={\pcgaqmut}{#1},
      },
      create on use/cgs-d-w-only/.style={
        create col/copy column from table={\cgsqmdwonlyt}{#1},
      },
      create on use/cgs-d-full-no-mls/.style={
        create col/copy column from table={\cgsqmdfullnomlst}{#1},
      },
      create on use/cgs-d/.style={
        create col/copy column from table={\cgsqmdt}{#1},
      },
      create on use/cgs-u-w-only/.style={
        create col/copy column from table={\cgsqmuwonlyt}{#1},
      },
      create on use/cgs-u-full-no-mls/.style={
        create col/copy column from table={\cgsqmufullnomlst}{#1},
      },
      create on use/cgs-u/.style={
        create col/copy column from table={\cgsqmut}{#1},
      },
      columns={
        pcga-d,
        cgs-d-w-only,
        cgs-d-full-no-mls,
        cgs-d,
        cgs-u-w-only,
        cgs-u-full-no-mls,
        cgs-u,
        pcga-u
      },
    },
  }%
  \pgfplotstablenew
  [
    combine tables column=mono,
  ]
  {\pgfplotstablegetrowsof\cgsqmut}
  \mono
  \pgfplotstablenew
  [
    combine tables column=poly,
  ]
  {\pgfplotstablegetrowsof\cgsqmut}
  \poly
  \pgfplotstablenew
  [
    combine tables column=overall,
  ]
  {\pgfplotstablegetrowsof\cgsqmut}
  \overall
  \subcaptionbox{Monocrystalline}{%
    \pgfplotstabletypeset
    [
      wafer scores,
      every row 0 column cgs-d-full-no-mls/.style={
        highlight cell,
      },
      every row 1 column pcga-d/.style={
        highlight cell,
      },
      every row 2 column cgs-d-full-no-mls/.style={
        highlight cell,
      },
      every row 2 column cgs-d/.style={
        highlight cell,
      },
      every row 3 column cgs-d-full-no-mls/.style={
        highlight cell,
      },
      every row 0 column cgs-u-full-no-mls/.style={
        highlight cell,
      },
      every row 1 column pcga-u/.style={
        highlight cell,
      },
      every row 2 column pcga-u/.style={
        highlight cell,
      },
      every row 3 column pcga-u/.style={
        highlight cell,
      },
    ]
    \mono
  }%
  \par
  \medskip
  \subcaptionbox{Polycrystalline}{%
    \pgfplotstabletypeset
    [
      wafer scores,
      every row 0 column cgs-d-full-no-mls/.style={
        highlight cell,
      },
      every row 1 column cgs-d/.style={
        highlight cell,
      },
      every row 2 column cgs-d/.style={
        highlight cell,
      },
      every row 3 column cgs-d/.style={
        highlight cell,
      },
      every row 0 column cgs-u/.style={
        highlight cell,
      },
      every row 1 column cgs-u/.style={
        highlight cell,
      },
      every row 2 column cgs-u/.style={
        highlight cell,
      },
      every row 3 column cgs-u/.style={
        highlight cell,
      },
    ]
    \poly
  }%
  \par
  \medskip
  \subcaptionbox{Overall}{%
    \pgfplotstabletypeset
    [
      wafer scores,
      every row 0 column cgs-d-full-no-mls/.style={
        highlight cell,
      },
      every row 1 column cgs-d/.style={
        highlight cell,
      },
      every row 2 column cgs-d-full-no-mls/.style={
        highlight cell,
      },
      every row 2 column cgs-d/.style={
        highlight cell,
      },
      every row 3 column cgs-d-full-no-mls/.style={
        highlight cell,
      },
      every row 0 column cgs-u-full-no-mls/.style={
        highlight cell,
      },
      every row 1 column pcga-u/.style={
        highlight cell,
      },
      every row 2 column cgs-u/.style={
        highlight cell,
      },
      every row 3 column cgs-u/.style={
        highlight cell,
      },
    ]
    \overall
  }%
\end{table*}

Pixelwise scores for the simplified masks of both methods are given in
\cref{tab:pixelwise-qm}. For monocrystalline \gls{PV}
modules, \gls{PGA} generally achieves higher scores. However, highest scores
are achieved only for images for which the lens distortion has been removed. The
proposed method fails to segment a row of cells in a solar module resulting in a
lower recall. However, for polycrystalline \gls{PV} modules, the proposed method
consistently outperforms \gls{PGA}. In the overall score, the proposed method
also outperforms the best-case evaluation for \gls{PGA} on undistorted images.
However, \gls{PGA} has highest recall, which is due to the lower number of
parameters of \gls{PGA}.

\paragraph{Weighted Jaccard Index}

\begin{figure*}[tp]
  \tikzsetnextfilename{jaccard-scores-boxplots}
  \includegraphics[width=\linewidth]{figures/jaccard-scores-boxplots}
  \caption{Boxplots of Jaccard scores for the three evaluated modalities. The
    Jaccard scores are computed against hand-labeled ground truth masks. In
    \subref{fig:solar-cells-jaccard-scores}, the scores are computed for the
    individual solar cells. In \subref{fig:solar-modules-jaccard-scores}, the
    scores are evaluated against the whole solar modules. The two left-most
    groups in each figure correspond to boxplots with respect to different
    solar wafers. Whereas the right-most group summarizes the performance of
    both solar wafer types combined.}
  \label{fig:el-pv-segm-accuracy}
\end{figure*}


The Jaccard scores summarized as boxplots in \cref{fig:el-pv-segm-accuracy}
support the pixelwise classification scores, showing that the proposed method
is more accurate than \gls{PGA}. The latter, however, is slightly more robust.
For complete modules, the considerable spread of \Gls{PGA} is partially
attributed to one major outlier. Overall, the proposed segmentation pipeline is
highly accurate. Particularly once a cell is detected, the cell outline is
accurately and robustly segmented.

\begin{additionenv}[label=a:ablation,ref={c:unwarping,c:ablation2},break]
\subsubsection{Ablation Study}
\label{sec:ablation}

We ablate the lens distortion parameters and the post~hoc application of affine
\gls{MLS} to investigate their effect on the accuracy and the success rate of
the segmentation process. The ablation is performed both on original (\ie,
distorted) \gls{EL} images and undistorted ones.

\paragraph{Distorted vs\@. Undistorted \gls{EL} Images}

For the ablation study, we consider two main cases. In the undistorted case,
both reference and predicted masks are unwarped using estimated lens distortion
parameters. Then, quadrilaterals are fitted to individual cell masks to allow a
comparison against \gls{PGA} which always yields such quadrilateral cell masks.
For a fair comparison, \gls{PGA} is also applied to undistorted images.

In the distorted case, however, the comparison is performed in the original
image space. Since the proposed method yields a curved grid after applying the
inverse of lens distortion, we synthesize a regular grid from backwarped cell
masks. Specifically, we extract the contours of estimated solar cell masks to
obtain the coordinates of the quadrilateral in the unwarped image, and then
apply the inverse of estimated geometric transformations to rectangle
coordinates. Afterwards, we fit lines to each side of the backwarped
quadrilaterals along grid rows and columns. From their intersections we finally
obtain the corner coordinates of each solar cells in the distorted image which
we can use for comparison against distorted \gls{PGA} results.

\paragraph{Parameterization}

First, we reduce the lens distortion model to a single radial distortion
parameter~\(\omega\) and assume both square aspect ratio (\ie, \(s_x=1\)) and
the center of distortion to be located in the image center. During optimization,
these two parameters are kept constant. In this experiment, we also do not
correct the curve grid using affine \gls{MLS}. The comparison against \gls{PGA}
shows that such a simplistic lens model is still more accurate than \gls{PGA}
both in the distorted and undistorted cases (\cf, \cref{tab:rmse}). However,
while the precision is high, the recall and therefore the \(F_1\)~score drops
considerably (\cf, \cref{tab:pixelwise-qm}). The reason for this is that such a
lens parametrization is too rigid. As a consequence, this weakens the grid
detection: correctly detected curves are erroneously discarded because of
inaccuracies of the lens model with only a single parameter~\(\omega\) instead
of four parameters (\(\omega,s_x,\vec c\)).

For the next comparison, we increase the number of degrees of freedom. Both the
distortion aspect ratio~\(s_x\) and the center of distortion~\(\vec c\) are
refined in addition to the radial distortion parameter~\(\omega\). Curve grid
correction via affine \gls{MLS} is again omitted. This parametrization achieves
much improved \gls{RMSE} and segmentation success rates.

Finally, we use the full parametrization, \ie, we refine all lens distortion
parameters~\((\omega,s_x,\vec c)\) and apply post~hoc correction via
affine~\gls{MLS}. This model is denoted as \textsl{w/~MLS}.

\paragraph{Discussion}

We summarize the results of the ablation study in
\cref{tab:rmse,tab:pixelwise-qm}. Here, \textsl{w/~MLS} denotes the full model that
includes the correction step via affine \gls{MLS}. The full model with post~hoc
affine \gls{MLS} grid correction performs in many instances best. However,
applying \gls{MLS} is not always beneficial. Particularly, for monocrystalline
\gls{PV} modules, grid correction does not always improve the results.

We conclude that the proposed joint lens model estimation with full
parametrization and grid detection is essential for robustness and accuracy of
the segmentation. Since the subsequent grid correction using affine \gls{MLS}
only marginally improves the results, its application can be seen as optional.

\end{additionenv}

\subsubsection{Segmentation Performance with Exact Cell Masks}

To allow an exact comparison of the segmentation results to the ground truth,
we inverse-warp the estimated solar cell masks back to the original image space
by using the determined perspective projection and lens distortion parameters.
This way, the estimated solar module masks will as exactly as possible overlay
the hand-labeled ground truth masks.

\paragraph{Pixelwise Classification}

\Cref{tab:pixelwise-em} summarizes the pixelwise classification scores for the
exact masks estimated using the proposed method. The method is more robust on
polycrystalline \gls{PV} modules than on monocrystalline modules. However, for
both module types, the method achieves a very high overall accuracy beyond
\SI{97}{\percent} for all metrics.
Investigation of failure cases for monocrystalline modules reveals difficulties
on cells where large gaps coincide with cell cracks and ragged edges.

\begin{table}[tbp]
  \centering
  \caption{Pixelwise classification scores for exact masks estimated using the
  \label{tab:pixelwise-em}
  proposed approach}
  \pgfplotstableset{
    col sep=comma,
    search path={data/pixelwise_scores},
  }%
  \pgfplotstableread{cgs_em_d.csv}\cgsemd
  \pgfplotstabletranspose[transposed scores]\cgsemdt\cgsemd
  \pgfplotstabletypeset[variant scores]\cgsemdt%
\end{table}



\paragraph{Weighted Jaccard Index}


Jaccard scores for exact masks are given in \cref{fig:jaccard-scores-em}. The
scores confirm the results of the pixelwise metrics. Notably, the \gls{IQR} of
individual cells has a very small spread, which indicates a highly consistent
segmentation. The \gls{IQR} of whole modules is slightly larger. This is,
however, not surprising since the boxplots summarize the joint segmentation
scores across multiple modules.

\begin{figure}[tp]
  \centering
  \tikzsetnextfilename{cgs-em-jaccard-scores-boxplots}%
  \includegraphics[width=\linewidth]{figures/jaccard-scores-boxplots-em}%
  \caption{Boxplots of Jaccard scores for the proposed approach}
  \label{fig:jaccard-scores-em}
\end{figure}

\subsection{Qualitative Results}
\label{sec:qualitative_results}

\begin{figure*}[tp]
  \centering
  \tikzsetnextfilename{segmentation-qualitative-results}%
  \input{figures/qualitative-results}
  \caption{Qualitative segmentation results of four test images depicting the
  estimated curve grid superimposed over the contrast-normalized input \gls{EL}
image. For visualization purposes, the original \gls{EL} images were cropped.}
  \label{fig:el-pv-qual-results}
\end{figure*}

\Cref{fig:el-pv-qual-results} shows the qualitative results of the segmentation
pipeline on four test images. The two results in the left column are computed on
monocrystalline modules, the two results in the right column on polycrystalline
modules. The estimated solar module curve grids are highly accurate. Even in
presence of complex texture intrinsic to the material, the accuracy of the
predicted solar module curve grid is not affected.

\subsection{Runtime Evaluation}
\label{sec:runtime-evaluation}

\begin{figure}[tb]
  \centering
  \tikzsetnextfilename{individual-processing-times}
  \includegraphics[width=\linewidth]{figures/individual-processing-times}
  \caption{Average time taken by individual steps of the segmentation pipeline,
  in seconds. The error bars denote the upper range of the standard deviation.}
  \label{fig:el-pv-segm-runtime-std}
\end{figure}

\begin{figure}[tb]
  \centering
  \tikzsetnextfilename{cumulative-processing-times}
  \includegraphics[width=\linewidth]{figures/cumulative-processing-times}
  \caption{Relative contribution of the average processing time for individual
  pipeline steps to the overall runtime with respect to different solar module
types and both types combined.}
  \label{fig:el-pv-segm-runtime}
\end{figure}

\Cref{fig:el-pv-segm-runtime-std} breaks down the average time taken by the
individual steps of the segmentation pipeline. \Cref{fig:el-pv-segm-runtime}
summarizes the contribution of individual pipeline steps to the overall
processing time for all 44~images. The timings were obtained on a consumer
system with an Intel i7-3770K CPU clocked at~\SI{3.50}{\giga\hertz} and
\SI{32}{\giga\byte} of~RAM. The first three stages of the segmentation pipeline
are implemented in \cpp\ whereas the last stage (except for \gls{MLS} image
deformation) is implemented in Python.

\phantomsection
\label{p:runtime-not-optimized}

For this benchmark, \gls{EL} images were processed sequentially running only on
the CPU. Note, however, that the implementation was not optimized in terms of
the runtime and only parts of the pipeline utilize all available CPU cores. To
this end, additional speedup can be achieved by running parts of the pipeline in
parallel or even on a GPU.

On average, it takes \SI{1}{\minute} and~\SI{6}{\second} to segment all solar
cells in a high resolution \gls{EL} image (\cf \cref{fig:el-pv-segm-runtime}).
Preprocessing is computationally most expensive, curve and cell extraction are
on average cheapest. The standard deviation of the model estimation step is
highest  (see \cref{fig:el-pv-segm-runtime-std}), which is mostly due to
dependency upon the total number of ridge edges and the number of resulting
curves combined with the probabilistic nature of \gls{LO-RANSAC}.

Interestingly, processing \gls{EL} images of monocrystalline solar modules takes
slightly longer on average than processing polycrystalline solar modules. This
is due to large gaps between ridges caused by cut-off corners that produce many
disconnected curve segments which must be merged first. Conversely, curve
segments in polycrystalline solar modules are closer, which makes it more likely
that several curve segments are combined early on.

\addition[label=a:manual-processing,ref=c:runtime]{An average processing time of
\SI{1}{\minute} and~\SI{6}{\second} is substantially faster than manual
processing, which takes at least several minutes.}
\addition[label=a:field-measurements,ref=c:runtime]{
For on-site \gls{EL} measurements with in-situ imaging of \gls{PV} modules, the processing times must be further optimized, likely by at least a factor of ten.
However, in other imaging environments, for example material testing
laboratories, the runtime is fully sufficient, given that the handling of each module for \gls{EL}
measurements and the performance evaluation impose much more severe scheduling bottlenecks.}

\subsection{Limitations}
\label{sec:el-pv-segm-limits}

Mounts that hold \Gls{PV} modules may cause spurious ridge edges. Early stages
of the segmentation focus on ridges without analyzing the whole image content,
which may occasionally lead to spurious edges and eventually to an incorrect
segmentation. Therefore, automatic image cropping prior to \gls{PV} module
segmentation could help reduce segmentation failures due to visible mounts.

While the algorithm is able to process disconnected (dark) cells, rows or
columns with more than \SI{50}{\percent} of disconnected cells pose a difficulty
in correctly detecting the grid due to insufficient edge information. However,
we observed that also human experts have problems to determine the contours
under such circumstances.

We also observed that smooth edges can result in segmentation failures. This is
because the stickness of smooth edges is weak and may completely fade away
after non-maximum suppression. This problem is also related to situations where
the inter-cell borders are exceptionally wide.  In such cases, it is necessary
to adjust the parameters of the ridgeness filter and the proximity of the
tensor voting.

\section{Conclusions}
\label{sec:el-pv-seg-conclusions}

In this work, we presented a fully automatic segmentation method for precise
extraction of solar cells from high resolution \gls{EL} images. The proposed
segmentation is robust to underexposure, and works robustly in presence of
severe defects on solar cells. This can be attributed to the proposed
preprocessing and the ridgeness filtering, coupled with tensor voting to
robustly determine the inter-cell borders and busbars.  The segmentation is
highly accurate, which allows to use its output for further inspection tasks,
such as automatic classification of defective solar cells and the prediction of
power loss.

We evaluated the segmentation with the Jaccard index on eight different
\gls{PV} modules consisting of \num{408} hand-labeled solar cells. The proposed
approach is able to segment solar cells with an accuracy
of~\SI{97.79682179077072}{\percent}. With respect to classification
performance, the segmentation pipeline reaches an~\(F_1\)~score
of~\SI{97.61704495045228}{\percent}.

Additionally, we compared the proposed method against the \gls{PV} module
detection approach by \citet{Sovetkin2019}, which is slightly more robust but
less accurate than our method.
\addition[label=a:decoupled-inferior,ref=c:fov-end-to-end]{The comparison also
shows that our joint lens distortion estimation and grid detection approach
achieves a higher accuracy than a method that decouples both steps.}

Beyond the proposed applications, the method can serve as a starting point for
bootstrapping deep learning architectures that could be trained end-to-end to
directly segment the solar cells. Future work may include to investigate the
required adaptations and geometric relaxations for using use the method not only
in manufacturing setting but also in the field.
\addition[label=a:end-to-end-grid-detection,ref=c:fov-end-to-end]
{Such relaxations could be achieved, for instance, by performing the grid
detection end-to-end using a \gls{CNN}.}

\addition[label=a:other-domains,ref=c:other-domains]{Given that grid structure
  is pervasive in many different problem domains, the proposed joint lens
  estimation and grid identification may also find other application fields,
  for example the detection of \gls{PV} modules in aerial imagery of solar
  power plants, building facade segmentation, and checkerboard pattern
  detection for camera calibration.}

\begin{acknowledgements}
This work was funded by Energy Campus Nuremberg~(EnCN) and partially
supported by the Research Training Group~1773 \enquote{Heterogeneous Image
Systems} funded by the German Research Foundation~(DFG).
\end{acknowledgements}

\bibliographystyle{spbasic}
\bibliography{references}

\end{document}

%% file: pgfstyles.tex
\pgfmathsetlengthmacro{\legendimageysep}{1pt}
\pgfmathsetlengthmacro{\crossrefyshift}{.5ex}

\pgfplotsset{
  enlarge legend image/.code={
    \node[fit=(current bounding box), inner xsep=0pt,
          inner ysep=#1, outer sep=0pt] {};
  },
  legend image code/.append style={
    /pgfplots/enlarge legend image=\legendimageysep,
  },
  every crossref picture/.style={
    baseline,
    yshift=\crossrefyshift,
  },
}

\pgfplotsset{
  every axis legend/.append style={
    trim left=(current bounding box.west),
    trim right=(current bounding box.east)
  },
  every legend to name picture/.append style={
    trim left=(current bounding box.west),
    trim right=(current bounding box.east)
  },
}

\pgfplotsset{table boxplot/.code args={row #1 of #2}{
  \pgfplotstableread{#2}\module

  \pgfplotstablegetelem{#1}{lower whisker}\of{\module}
  \edef\lowerw{\pgfplotsretval}

  \pgfplotstablegetelem{#1}{lower quartile}\of{\module}
  \edef\lowerq{\pgfplotsretval}

  \pgfplotstablegetelem{#1}{median}\of{\module}
  \edef\median{\pgfplotsretval}

  \pgfplotstablegetelem{#1}{upper quartile}\of{\module}
  \edef\upperq{\pgfplotsretval}

  \pgfplotstablegetelem{#1}{upper whisker}\of{\module}
  \edef\upperw{\pgfplotsretval}

  \pgfplotsset{boxplot prepared={
      lower whisker=\lowerw,
      lower quartile=\lowerq,
      median=\median,
      upper quartile=\upperq,
      upper whisker=\upperw,
  }}
}}

\pgfplotsset{
  group boxplots/.style={
    boxplot={
      draw position={1/(#1+1)+floor(\plotnumofactualtype/#1)+1/(#1+1)*mod(\plotnumofactualtype,#1)},
    },
  }
}

\tikzset{
  per module colors/.style={pattern=crosshatch,draw=GnBu-G!50!black,pattern
  color=GnBu-E!25!black},
  per cell colors/.style={pattern=north west lines,draw=GnBu-M,pattern
  color=GnBu-H!50},
}

\pgfplotsset{
  boxplot/draw/whisker/.code 2 args={
    \draw [/pgfplots/boxplot/every whisker/.try,thin]
      (boxplot cs:#1) -- (boxplot cs:#2);
    \draw [thick,/pgfplots/boxplot/every whisker/.try,solid]
      (boxplot whisker cs:#2,0) -- (boxplot whisker cs:#2,1);
  },
  boxplot/every box/.append style={black},
  boxplot/every median/.append style={very thick},
  boxplot/every whisker/.append style={gray},
  boxplot/every box/.append style={solid},
  boxplot/every median/.append style={solid},
  boxplot/every whisker/.append style={solid},
}

\colorlet{major grid}{Paired-H}
\colorlet{busbars}{Paired-A}
\colorlet{reference grid}{Paired-F}

\tikzset{
  module/.style={
    mark size=1pt,
  },
  major grid/.style={
    color=major grid,
    semithick,
  },
  convex hull/.style={
    major grid,
    thick,
  },
  major grid intersections only/.style={
    Paired-H,
    fill=.!25,
    mark=*,
    mark options={solid},
  },
  major grid intersections/.style={
    major grid intersections only,
    only marks,
  },
  busbars/.style={
    densely dashed,
    color=busbars,
  },
  minor grid intersections only/.style={
    Paired-B,
    fill=.!25,
    mark=*,
    mark options={solid},
  },
  minor grid intersections/.style={
    minor grid intersections only,
    only marks,
  },
  reference grid/.style={
    thick,
    color=reference grid,
  },
}

\tikzset{
  module lines/.style={
    /pgfplots/.cd,
    colormap/Paired,
    /tikz/.cd,
    every path/.style={
      line cap=round,
      mark=*,
      mark size=0.75pt,
      mark options={fill=.!50, line width=0.5pt},
      thick,
    },
    every path/.append code={
      \pgfmathsetmacro\currentindex{mod(\cmindex+1,\cmcount}
      \global\cmindex=\currentindex
      \tikzset{index of colormap=\cmindex}
    },
  },
  module lines/.append code={
    \newcount\cmindex\cmindex=-1
    \newcount\cmcount
    \pgfkeysgetvalue{/pgfplots/colormap name}\colormap
    \pgfmathsetmacro\cmcount{\pgfplotscolormapsizeof{\colormap}}
  },
}

\tikzset{
  parabola/.style={
    thick,
  },
  parabola/vertical/.style={
    parabola,
    Paired-A,
  },
  parabola/horizontal/.style={
    parabola,
    Paired-C,
  },
}

\makeatletter
\pgfplotsset{boxplot/no outliers/.code={
  \def\pgfplotsplothandlerboxplot@outlier{}%
}}
\makeatother

\def\decomposetikzcoordinate#1,#2{%
  \pgfpointxy{#1}{#2}%
}

\pgfplotsset{
  boxplot legend cs/.code={
    \tikzdeclarecoordinatesystem{boxplot box}{%
      \decomposetikzcoordinate##1\relax%
    }%
  },
  boxplot legend/.style={
    anchor=north,
    legend image code/.code={%
      \begin{scope}
        [
          x=1mm,
          y=1mm,
          shift={(0,-1)},
          yshift=-0.1em,
          scale=3,
          rotate=90,
          /pgfplots/boxplot/.cd,
          lower quartile=0,
          upper quartile=1,
          median=0.5,
        ]
        \pgfplotsset{boxplot legend cs}
        \pgfplotsset{boxplot/draw/box}
        \pgfplotsset{boxplot/draw/median}
      \end{scope}
    }
  }
}

\tikzset{
  subfigures/.code={
    \setcounter{subfigure}{0}
  }
}

\pgfplotsset{
  boxplot/solar cells/.style={
    boxplot/every box/.append style={Paired-A,pattern=crosshatch,draw=.!50!black,pattern color=.!90!black},
    boxplot/every median/.append style={Paired-B,.!50!black},
  },
  boxplot/modules/.style={
    boxplot/every box/.append style={Paired-C,pattern=north west lines,draw=.!50!black,pattern color=.!90!black},
    boxplot/every median/.append style={Paired-D,.!70!black},
  },
}

\tikzset{
  boxplot/pcga/.style={
    Set2-A,.!70!black,pattern color=.!60,
  },
}

\pgfplotsset{
  boxplot/pcga-qm-d/.style={
    boxplot/every box/.append style={boxplot/pcga,pattern=crosshatch dots},
    boxplot/every median/.append style={boxplot/pcga,.!60!black},
  },
  boxplot/pcga-qm-u/.style={
    boxplot/every box/.append style={boxplot/pcga,pattern=north east lines},
    boxplot/every median/.append style={boxplot/pcga,.!60!black},
  },
  boxplot/cgs-qm-d/.style={
    boxplot/every box/.append style={Set1-E,pattern=north west lines,draw=.!75!black,pattern color=.!50},
    boxplot/every median/.append style={Set1-E,.!70!black},
  },
}

\pgfplotstableset{
  make cell content bold/.style={
    /pgfplots/table/@cell content/.add={\bfseries\boldmath}{},
  },
  bold cell content/.style={
    postproc cell content/.append style={make cell content bold},
  }
}

\pgfplotstableset{
  create on use/metric/.style={
    create col/set list={Precision,Recall,\(F_1\) score,Accuracy},
  },
  create on use/type/.style={
    create col/set list={mono,poly,overall},
  },
  columns/type/.style={
    string type,
    column name=,
  },
  percent/.style={
    postproc cell content/.append code={
      \ifnum0=\pgfplotstablepartno
        \pgfkeysalso{@cell content/.add={}{\,\si{\percent}}}%
      \fi
    },
  },
  score/.style={
    percent,
    /pgf/number format/fixed,
    /pgf/number format/fixed zerofill,
    /pgf/number format/precision=2,
  },
  variant scores/.style={
    begin table=\begin{tabularx}{\columnwidth},
    end table=\end{tabularx},
    every head row/.style={
      before row=\toprule,
      after row=\midrule,
    },
    every last row/.style={
      after row=\bottomrule,
    },
    columns/metric/.style={
      string type,
      column name=Metric,
      column type=X,
    },
    columns/mono/.style={
      column name=Monocrystalline,
      score,
    },
    columns/poly/.style={
      column name=Polycrystalline,
      score,
    },
    columns/overall/.style={
      column name=Overall,
      score,
    },
    columns={metric,mono,poly,overall},
  },
  transposed scores/.style={
    columns={type,precision,recall,f1-score,accuracy},
    input colnames to=,
    colnames from=type,
  },
  ablation header/.style={
    before row={
      \toprule
      &
      \multicolumn{4}{c}{Distorted}
      &
      \multicolumn{4}{c}{Undistorted}
      \\
      \cmidrule(lr){2-5}
      \cmidrule(l){6-9}
      &
      \multirow{2}{*}{%
        \begin{tabular}{@{}c@{}}
          PGA
        \end{tabular}}%
      &
      \multicolumn{3}{c}{Proposed}
      &
      \multirow{2}{*}{%
        \begin{tabular}{@{}c@{}}
          PGA
        \end{tabular}}%
      &
      \multicolumn{3}{c}{Proposed}
      \\
      \cmidrule(r){3-5}
      \cmidrule(l){7-9}
    },
    after row=\midrule,
  },
  wafer scores/.style={
    begin table=\begin{tabularx}{\linewidth},
    end table=\end{tabularx},
    every head row/.style={
      ablation header,
    },
    every last row/.style={
      after row=\bottomrule,
    },
    columns/metric/.style={
      string type,
      column name=Metric,
      column type=l,
    },
    columns/pcga-d/.style={
      column name=\empty,
      column type=Y,
      score,
    },
    columns/pcga-u/.style={
      column name=\empty,
      column type=Y,
      score,
    },
    columns/cgs-d/.style={
      column name=w/~MLS,
      column type=Y,
      score,
    },
    columns/cgs-u-w-only/.style={
      column name=\(\omega\),
      column type=Y,
      score,
    },
    columns/cgs-d-w-only/.style={
      column name=\(\omega\),
      column type=Y,
      score,
    },
    columns/cgs-u-full-no-mls/.style={
      column name={\((\omega,s_x,\vec c)\)},
      column type=Y,
      score,
    },
    columns/cgs-d-full-no-mls/.style={
      column name={\((\omega,s_x,\vec c)\)},
      column type=Y,
      score,
    },
    columns/cgs-u/.style={
      column name=w/~MLS,
      column type=Y,
      score,
    },
    columns={
      metric,
      pcga-d,
      cgs-d-w-only,
      cgs-d-full-no-mls,
      cgs-d,
      pcga-u,
      cgs-u-w-only,
      cgs-u-full-no-mls,
      cgs-u
    },
  },
}

\tikzset{mask/module cell/.style={
  fill,
  fill opacity=0.65,
}}

\pgfplotsset{
  colormap/Set2,
  colormap/Dark2,
  colormap/Paired,
}

%% file: figures/cell-extraction-overview.tex
\begin{tikzpicture}
  [
    tight background,
  ]

  \pgfdeclareimage[width=.1\linewidth]{cell_1_1}{images/160429_AZT_ITS_00137__t5000ms_i5625mA_D_PM010022_cells/160429_AZT_ITS_00137__t5000ms_i5625mA_D_PM010022_1_1}
  \pgfdeclareimage[width=.1\linewidth]{cell_1_2}{images/160429_AZT_ITS_00137__t5000ms_i5625mA_D_PM010022_cells/160429_AZT_ITS_00137__t5000ms_i5625mA_D_PM010022_1_2}
  \pgfdeclareimage[width=.1\linewidth]{cell_1_3}{images/160429_AZT_ITS_00137__t5000ms_i5625mA_D_PM010022_cells/160429_AZT_ITS_00137__t5000ms_i5625mA_D_PM010022_1_3}

  \pgfdeclareimage[width=.1\linewidth]{cell_2_1}{images/160429_AZT_ITS_00137__t5000ms_i5625mA_D_PM010022_cells/160429_AZT_ITS_00137__t5000ms_i5625mA_D_PM010022_2_1}
  \pgfdeclareimage[width=.1\linewidth]{cell_2_2}{images/160429_AZT_ITS_00137__t5000ms_i5625mA_D_PM010022_cells/160429_AZT_ITS_00137__t5000ms_i5625mA_D_PM010022_2_2}
  \pgfdeclareimage[width=.1\linewidth]{cell_2_3}{images/160429_AZT_ITS_00137__t5000ms_i5625mA_D_PM010022_cells/160429_AZT_ITS_00137__t5000ms_i5625mA_D_PM010022_2_3}

  \begin{scope}
    [
      my spy/.style={
        spy scope={
          every spy in node/.style={draw,ultra thick},
          every spy on node/.style={draw,thick},
          #1,
          spy connection path={\draw[thick] (tikzspyonnode) -- (tikzspyinnode);},
        }
      },
      my spy={Dark2-A, circle, magnification=5, size=2.65cm, connect spies, very thick},
    ]
  \node[outer sep=0pt,inner sep=0pt,anchor=south west] (solar module) {%
    \setkeys{Gin}{width=.5\linewidth}%
    \input{figures/intermediate/160429_AZT_ITS_00137__t5000ms_i5625mA_D_PM010022_grid}%
  };

  \node[xshift=1em,label={[rotate=90,anchor=south]left:Row 1},right=.75 of solar module]
    (cell-1-1) {\pgfuseimage{cell_1_1}};
  \node[anchor=north west] (cell-1-2) at ([xshift=1em, yshift=-1em]cell-1-1.north west)
    (cell-1-2) {\pgfuseimage{cell_1_2}};
  \node[anchor=north west] (cell-1-3) at ([xshift=1em, yshift=-1em]cell-1-2.north west)
    (cell-1-3) {\pgfuseimage{cell_1_3}};
  \node[anchor=north, font=\LARGE] at ([yshift=1em]cell-1-3.south) {\( \ddots \) };

  \node[label={[rotate=90,anchor=south]left:Row 2},right=1.5 of cell-1-1] (cell-2-1)
    {\pgfuseimage{cell_2_1}};
  \node[anchor=north west] (cell-2-2) at ([xshift=1em, yshift=-1em]cell-2-1.north west)
    (cell-2-2) {\pgfuseimage{cell_2_2}};
  \node[anchor=north west] (cell-2-3) at ([xshift=1em, yshift=-1em]cell-2-2.north west)
    (cell-2-3) {\pgfuseimage{cell_2_3}};
  \node[anchor=north, font=\LARGE] at ([yshift=1em]cell-2-3.south) {\( \ddots \) };

  \begin{scope}[x={(solar module.south east)},y={(solar module.north west)}]
    \coordinate (first spy on) at (0.035,0.875);
    \coordinate (first spy at) at (0.3,0.55);

    \coordinate (second spy on) at (0.96,0.86);
    \coordinate (second spy at) at (0.7,0.55);

    \spy on (first spy on) in node at (first spy at);
    \spy on (second spy on) in node at (second spy at);

  \end{scope}

  \end{scope}

  \begin{scope}[gray,thick,->]
    \draw (solar module.26) to [in=100, out=0] (cell-1-1.north);
    \draw (solar module.16) to [in=100, out=0] (cell-2-1.north);
  \end{scope}

  \node[right=.25 of cell-2-3.east, font=\LARGE] { \( \dots \) };

  \begin{scope}
    [
      inner sep=0pt,
      outer sep=0pt,
      text width=.5\linewidth,
    ]
    \node[below] (left caption) at (solar module.south) {%
      \subcaption{}%
      \label{fig:el-pv-curve-grid}%
    };

    \node[anchor=west] at (left caption.east) {%
      \subcaption{}%
      \label{fig:el-pv-extracted-cells}%
    };
  \end{scope}
\end{tikzpicture}%

%% file: figures/cell-extraction-pipeline.tex
\pgfdeclarelayer{foreground}
\pgfdeclarelayer{background}
\pgfsetlayers{background,main,foreground}

\colorlet{seg inner}{Paired-B!15}
\colorlet{seg outer}{Paired-B}
\colorlet{curve inner}{Paired-D!15}
\colorlet{curve outer}{Paired-D}
\colorlet{model inner}{Paired-J!15}
\colorlet{model outer}{Paired-J}
\colorlet{rectify inner}{Dark2-B!15}
\colorlet{rectify outer}{Dark2-B}

\def\stagewidth{3cm}

\begin{tikzpicture}
  [
    every path/.style={semithick},
    shorten >=1pt,
    node distance=1em,
    pstep simple/.style={draw=none,rectangle,align=center,
      minimum width=\stagewidth,
      inner sep=0.5em,
      text depth=0pt,
    },
    pstep seg/.style={pstep simple,seg outer,draw,fill=white},
    pstep curve/.style={pstep simple,curve outer,draw,fill=white},
    pstep model/.style={pstep simple,model outer,draw,fill=white},
    pstep rectify/.style={pstep simple,rectify outer,draw,fill=white},
  ]
  \node[pstep seg] (image) {Local ridge\\features};

  \node[pstep curve,right=1 of image.north east,anchor=north west]
    (subp) {Quadratic\\curves};

  \node[pstep model,right=1 of subp.north east,anchor=north west]
    (parameter estimate) {Coherent\\grid};

  \node[pstep rectify,right=1 of parameter estimate.north east,anchor=north west]
    (config est) {Solar cell\\ROIs};

  \begin{pgfonlayer}{background}
    \begin{scope}[inner sep=1.5ex, outer sep=0pt]
      \node[fill=seg inner,fit=(image)]
        (prep) {};
      \node[fill=curve inner,fit=(subp)]
        (curve ext) {};
      \node[fill=model inner,fit=(parameter estimate)] (model) {};
      \node[fill=rectify inner,fit=(config est)] (cell ext) {};
    \end{scope}
  \end{pgfonlayer}

  \begin{scope}[->]
    \path[seg outer] (image.east) edge (subp.west);
    \path[curve outer] (subp.east) edge (parameter estimate.west);
    \path[model outer] (parameter estimate.east) edge (config est.west);
  \end{scope}



  \begin{scope}[text width=\stagewidth]
    \node[below] at (prep.south) {%
      \subcaption{}%
      \label{fig:el-preprocessing}
    };
    \node[below] at (prep.south -| subp.south) {%
      \subcaption{}%
      \label{fig:el-curve-extraction}
    };
    \node[below] at (prep.south -| parameter estimate.south) {%
      \subcaption{}%
      \label{fig:el-curve-grid-model}
    };
    \node[below] at (prep.south -| config est.south) {%
      \subcaption{}%
      \label{fig:el-curve-grid-cell-extraction}
    };
  \end{scope}

  \begin{scope}[font=\bfseries,align=center,text depth=0pt]
    \node[above,seg outer] at (prep.north) {Preprocessing};
    \node[above,curve outer] at (curve ext.north) {Curve Extraction};
    \node[above,model outer] at (model.north) {Model Estimation};
    \node[above,rectify outer] at (cell ext.north) {Cell Extraction};
  \end{scope}
\end{tikzpicture}

%% file: figures/curves-heuristics.tex
\pgfdeclarelayer{background}%
\pgfsetlayers{background,main}%
\begin{tikzpicture}
[
  segment/.style={arrows={{Circle[length=4pt]}-{Circle[length=4pt]}},very thick},
  segments/.style={very thick,mark=*,mark size=(4pt-\pgflinewidth)*0.5},
  first/.style={Dark2-A},
  second/.style={Dark2-B},
  every path/.style={scale=1.25},
  declare function={
    angle=22.5;
    segment_anchor=angle+45;
  },
]
  \draw[densely dotted,Dark2-A] (0,0) -- +(angle:0.5) coordinate (A);
  \draw[segment,first] (A)
    node[anchor=-segment_anchor] {\(A\)}
    --
    +(angle:1.2)
    coordinate (B)
    node[anchor=-segment_anchor] {\(B\)}
    ;

  \draw[segment,second]
    ($(B)+(2.25,0)$) coordinate (C)
    node[anchor=180+segment_anchor] {\(B\mathrlap{'}\)}
    --
    +(-22.5:1) coordinate (D)
    node[anchor=180+segment_anchor] {\(A\mathrlap{'}\)}
    ;
  \draw[densely dotted,second] (D) -- +(-angle:.5) coordinate (E);

  \path[name path=BC] (B) -- (angle:5) coordinate (AB ext);
  \path[name path=CD] (C) -- +(-angle:-3) coordinate (CD ext);

  \path[name intersections={of=BC and CD,by=inter}];

  \draw[first,dashed] (B) -- (inter);
  \draw[second,dashed] (inter) -- (C);

  \colorlet{angle}{Dark2-E}

  \draw (D)
    let \p1 = ($(C)-(inter)$) in
    pic
    [
      draw=Dark2-C,
      pattern=north east lines,
      pattern color=Dark2-C,
      semithick,
      angle radius={0.3*veclen(\x1,\y1)}
    ]
    {angle=B--inter--C}
    ;

  \begin{pgfonlayer}{background}
    \draw (B)
      let \p1 = ($(inter)-(C)$) in
      pic [Dark2-B,draw=.,fill=.!25,angle radius={0.3*veclen(\x1,\y1)}]
        {angle=B--C--D}
      ;

    \draw (inter)
      let \p1 = ($(inter)-(B)$) in
      pic [Dark2-A,draw=.,fill=.!25,angle radius={0.3*veclen(\x1,\y1)}]
        {angle=A--B--C}
      ;
  \end{pgfonlayer}

  \draw[dashed] (B) -- (C);

  \begin{scope}[outer sep=.25ex]
    \node[Dark2-A,anchor=90+angle/2] at (B) {\(\alpha_1\)};
    \node[Dark2-B,anchor=90-angle/2] at (C) {\(\alpha_2\)};
    \node[anchor=south,Dark2-C] at (inter) {\(\alpha_3\)};
  \end{scope}

  \draw[mark=x,very thick,Dark2-H,scale=1.5] plot coordinates {(inter)};
\end{tikzpicture}%

%% file: figures/preprocessing.tex
\begin{tikzpicture}
  [
    subfigures,
    tight background,
  ]
  \graphicspath{{./images/intermediate/}}
  \setkeys{Gin}{trim=0.0bp 512.75bp 51.150000000000006bp 20.51bp}%
  \matrix
  [
    inner sep=0pt,
    matrix of nodes,
    column sep={1em,between borders},
    row sep={1em,between borders},
  ]
  {
    \subcaptionbox{\Gls{EL} image of a monocrystalline \gls{PV}
      module\label{fig:el-pv-seg-input}}
      {\includegraphics[clip]{160429_AZT_ITS_00137__t5000ms_i5625mA_D_PM010022_original}}%
    &
    \subcaptionbox{Background-equalized image\label{fig:el-pv-seg-bgeq-example}}
      {\includegraphics[clip]{160429_AZT_ITS_00137__t5000ms_i5625mA_D_PM010022_bgnormalized}}%
    \\
    \subcaptionbox{Ridgeness image \(R(\vec u)\) from the filter responses
      at multiple scales\label{fig:el-pv-seg-ridgeness}}
      {\includegraphics[clip]{160429_AZT_ITS_00137__t5000ms_i5625mA_D_PM010022_rigidness}}%
    &
    \subcaptionbox{Stickness of ridgeness contextually enhanced using tensor
    voting\label{fig:el-pv-seg-stickness}}
      {\includegraphics[clip]{160429_AZT_ITS_00137__t5000ms_i5625mA_D_PM010022_stickness}}%
    \\
    \subcaptionbox{Extracted line segments grouped by their curvature}
      {\input{figures/intermediate/160429_AZT_ITS_00137__t5000ms_i5625mA_D_PM010022_lines}}%
    &
    \subcaptionbox{Horizontal (\hcurvemarker) and vertical (\vcurvemarker)
      parabolic curves filtered using the intersection
    constraint\label{fig:preprocessing-curves}}
      {\input{figures/intermediate/160429_AZT_ITS_00137__t5000ms_i5625mA_D_PM010022_parabolas}}%
    \\
  };
\end{tikzpicture}%

%% file: figures/cell-segments.tex
\tikzset{
  segment sep/.style={densely dashed,semithick,white},
  every picture/.append style={tight background},
}%
\colorlet{color1}{Paired-A}%
\colorlet{color2}{Paired-G}%
\graphicspath{{./images/segments/}}%
\newsavebox\segmentsA
\newsavebox\segmentsB
\newsavebox\segmentsC
\sbox\segmentsA{%
  \tikzexternaldisable
  \input{figures/cell-segments-1-by-1}%
}%
\sbox\segmentsB{%
  \tikzexternaldisable
  \input{figures/cell-segments-3-by-1}%
}%
\sbox\segmentsC{%
  \tikzexternaldisable
  \input{figures/cell-segments-4-by-1}%
}%
\begin{tikzpicture}
  \matrix
  [
    inner sep=0pt,
    matrix of nodes,
    column sep={1em,between borders},
    row sep={1em,between borders},
  ]
  {
    \subcaptionbox{\label{fig:no-subdivisions}}{%
      \usebox\segmentsA
    }%
    &
    \subcaptionbox{\label{fig:three-segments}}{%
      \usebox\segmentsB
    }%
    &
    \subcaptionbox{\label{fig:four-segments}}{%
      \usebox\segmentsC
    }%
    \\
  };
\end{tikzpicture}%

%% file: figures/cell-segments-1-by-1.tex
\begin{tikzpicture}[external/optimize=false]
  \node
  [
   outer sep=0pt,
   inner sep=0pt,
   anchor=south west,
  ]
  (canvas) {\includegraphics{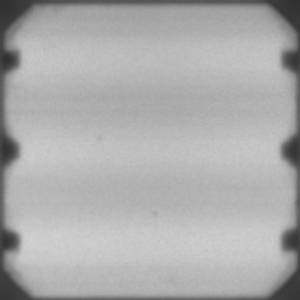}};

  \useasboundingbox (canvas.north west) rectangle (canvas.south east);

  \begin{scope}
    [
      x={(canvas.south east)},
      y={(canvas.north west)},
      yscale=-1, 
      shift={(0,-1)},
    ]

    \begin{scope}
      [
         <->,right,
         shift={(0.5,0)},
         gray!50!black,
      ]
      \draw (0,0) -- node {\(\Delta_1\)} (0,1);

    \end{scope}
  \end{scope}
\end{tikzpicture}%

%% file: figures/cell-segments-3-by-1.tex
\begin{tikzpicture}[external/optimize=false]
  \node
  [
   outer sep=0pt,
   inner sep=0pt,
   anchor=south west,
  ]
  (canvas) {\includegraphics{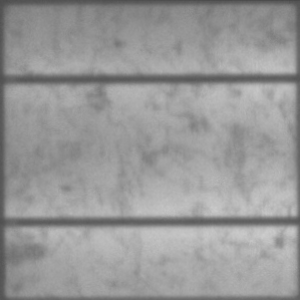}};

  \useasboundingbox (canvas.north west) rectangle (canvas.south east);

  \begin{scope}
    [
      x={(canvas.south east)},
      y={(canvas.north west)},
      yscale=-1, 
      shift={(0,-1)},
    ]

    \begin{scope}[opacity=0.4]
      \begin{scope}[fill=color1]
        \fill (0,0) rectangle (1,0.26) coordinate (top right);
        \fill (0,1-0.26) coordinate (bottom left)
               rectangle (1,1);
      \end{scope}

      \begin{scope}[fill=color2]
        \fill (bottom left) rectangle (top right);
      \end{scope}
    \end{scope}

    \begin{scope}[segment sep]
      \draw (0,0 |- top right) coordinate (top left) -- (top right);
      \draw (bottom left) -- (bottom left -| {(1,1)}) coordinate (bottom right);
    \end{scope}

    \begin{scope}
      [
         <->,right,
         shift={(0.5,0)},
      ]
      \draw[color1!50!black] (0,0) -- node {\(\Delta_1\)} ($(top left)+(0.5,0)$) coordinate (center top);
      \draw[color1!50!black] (0,1) -- node {\(\Delta_1\)} ($(bottom right)-(0.5,0)$) coordinate (center bottom);

      \draw[color2!50!black] (center top) -- node {\(\Delta_2\)} (center bottom);
    \end{scope}
  \end{scope}
\end{tikzpicture}%

%% file: figures/cell-segments-4-by-1.tex
\begin{tikzpicture}[external/optimize=false]
  \node
  [
   outer sep=0pt,
   inner sep=0pt,
   anchor=south west,
  ]
  (canvas) {\includegraphics{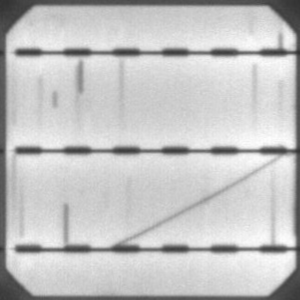}};

  \useasboundingbox (canvas.north west) rectangle (canvas.south east);

  \begin{scope}
    [
      x={(canvas.south east)},
      y={(canvas.north west)},
      yscale=-1, 
      shift={(0,-1)},
    ]

    \begin{scope}[opacity=0.5]
      \begin{scope}[fill=color1]
        \fill (0,0) rectangle (1,0.175) coordinate (top right);
        \fill (0,1-0.17) coordinate (bottom left) rectangle (1,1);
      \end{scope}

      \begin{scope}[fill=color2]
        \fill (0,0.175) rectangle (1,0.505) coordinate (center right);
        \fill (0,0.505) rectangle (1,1-0.17);
      \end{scope}
    \end{scope}

    \begin{scope}[segment sep]
      \draw (0,0 |- top right) coordinate (top left) -- (top right);
      \draw (bottom left |- center right) coordinate (center left) -- (center right);
      \draw (bottom left) -- (bottom left -| {(1,1)}) coordinate (bottom right);
    \end{scope}

    \begin{scope}
      [
         <->,right,
         shift={(0.5,0)},
      ]
      \draw[color1!55!black] (0,0) -- node {\(\Delta_1\)} ($(top left)+(0.5,0)$) coordinate (center 1 top);
      \draw[color1!55!black] (0,1) -- node {\(\Delta_1\)} ($(bottom right)-(0.5,0)$) coordinate (center 2 bottom);

      \draw[color2!55!black] (center 1 top) -- node {\(\Delta_2\)} (center 1 top |- center right) coordinate (center 1 bottom);
      \draw[color2!55!black] (center 1 bottom) -- node {\(\Delta_2\)} (center 1 bottom |- bottom right);
    \end{scope}
  \end{scope}
\end{tikzpicture}%

%% file: figures/mask-generation.tex
\begin{tikzpicture}
[
  tight background,
  trim left=(top left),
  trim right=(final cell br),
]

\graphicspath{{./images/mask/}}%

\pgfmathsetlengthmacro\profilesize{5mm}
\pgfmathsetlengthmacro\groupsep{1em-2*1ex}
\pgfmathsetlengthmacro\cellshift{2mm}
\pgfmathsetlengthmacro\plotwidth{%
  (\linewidth-\groupsep*4-3*\cellshift-\profilesize-14*1ex)/5
}

\begin{groupplot}
[
  group/horizontal sep=\groupsep,
  group/vertical sep=0pt,
  group/group size=6 by 2,
  group/group name=cell mask,
  xticklabel=\empty,
  yticklabel=\empty,
  axis on top,
  scale only axis,
  hide axis,
  enlargelimits={abs=1ex},
  width=\plotwidth,
  height=\plotwidth,
  plot graphics/xmin=0,
  plot graphics/ymin=0,
  plot graphics/xmax=300,
  plot graphics/ymax=300,
  xmin=0,
  ymin=0,
  xmax=300,
  ymax=300,
  table/search path={./data/mask},
  no marks,
  cell/.style={
    fill=gray!50,
    draw=gray,
  },
  cycle list/Set2,
  enlargelimits respects figure size=false,
]

\nextgroupplot[width=\plotwidth,height=\profilesize]

\nextgroupplot
[
  hide axis=false,
  height=\profilesize,
  width=\plotwidth,
  ymin=0,
  ymax=1,
  xmin=0,
  xmax=300,
  xtick={0,300},
  ytick={0,1},
  xmajorgrids,
  extra x ticks={6,295},
  table/y index=0,
  table/x expr=\coordindex,
  group/horizontal sep=\groupsep+3ex,
  every axis plot/.append style={
    thick,
  },
  legend columns=-1,
  legend entries={
    Mean profile,
    Thresholded profile
  },
  legend to name=cell profiles,
]
\addplot+ table {partial_2_v_profile.csv};
\addplot+[const plot] table {partial_2_v_projection.csv};

\nextgroupplot
[
  width=\profilesize,
  height=\profilesize,
  group/horizontal sep=0pt,
]

\nextgroupplot
[
  group/horizontal sep=\groupsep+1.5ex,
  width=\plotwidth,
  height=\profilesize,
]
\nextgroupplot[width=\plotwidth,height=\profilesize]
\nextgroupplot[width=\plotwidth,height=\profilesize]

\nextgroupplot
[
  clip=false,
]

\addplot[xshift=-3*\cellshift] graphics {2_1_1};
\addplot[xshift=-2*\cellshift] graphics {2_1_2};
\addplot[xshift=-1*\cellshift] graphics {2_1_3};

\coordinate[xshift=-3*\cellshift] (top left) at (axis cs:0,300);
\coordinate (bottom right) at (axis cs:300,0);

\addplot graphics {2_4_11};

\nextgroupplot
[
  xtick=\empty,
  ytick=\empty,
  grid=major,
  xtick={0,300},
  ytick={0,300},
  extra x ticks={6,295},
  extra y ticks={4,295},
  hide axis=false,
]

\addplot graphics {avg_image};

\nextgroupplot
[
  group/horizontal sep=0pt,
  hide axis=false,
  width=\profilesize,
  height=\plotwidth,
  xmin=0,
  xmax=1,
  ymin=0,
  ymax=300,
  xtick={0,1},
  ytick={0,300},
  ymajorgrids,
  extra y ticks={4,295},
  table/y expr=\coordindex,
  table/x index=0,
  every axis plot/.append style={
    thick,
  },
]
\addplot+ table {partial_2_h_profile.csv};
\addplot+[const plot] table {partial_2_h_projection.csv};

\nextgroupplot
[
]
\addplot[cell] table {partial_2_outer.csv}
  coordinate[pos=0.75] (aug cell bl0)
  coordinate[pos=0.50] (aug cell br0)
  coordinate[pos=0.25] (aug cell tr0)
  --
  cycle
  ;

\nextgroupplot[group/horizontal sep=0pt]
\pgfplotsinvokeforeach{0,...,5} {
  \addplot[cell] table {partial_2_extra_#1.csv}
    coordinate[pos=0.00] (extra geom #1 start)
    --
    cycle
    ;
}

\nextgroupplot[group/horizontal sep=0pt]
\addplot[cell] table {partial_2_cell.csv}
  coordinate[pos=0.50] (final cell bl0)
  coordinate[pos=0.00] (final cell br0)
  coordinate[pos=0.25] (final cell bottom)
  --
  cycle
  ;

\end{groupplot}

\node[xshift=-.5ex] at ($(cell mask c1r2.east)!.5!(cell mask c2r2.west)$) {%
  \(\cdots\)%
};

\coordinate (ext cell bl) at (top left |- bottom right);
\coordinate (ext cell br) at (bottom right);

\coordinate (avg cell bl) at (cell mask c2r2.south west);
\coordinate (avg cell br) at (cell mask c3r2.south east);

\coordinate (aug cell bl) at (aug cell bl0 |- aug cell br0);
\coordinate (aug cell br) at (aug cell br0 -| aug cell tr0);

\coordinate (final cell bl) at (final cell bl0 |- final cell bottom);
\coordinate (final cell br) at (final cell bl -| final cell br0);

\coordinate (extra geom bl) at (extra geom 3 start |- final cell bl);
\coordinate (extra geom br) at (extra geom 0 start |- extra geom bl);

\begin{scope}
[
  inner xsep=0pt,
  align=center,
  every node/.append style={%
    below,
  },
]

\coordinate (C11) at (cell mask c6r2.outer south -| cell mask c1r2.south);
\coordinate (C1) at (C11 -| cell mask c1r2.south);
\coordinate (C2) at (C11 -| cell mask c2r2.south);
\coordinate (C3) at (C11 -| cell mask c4r2.south);
\coordinate (C4) at (C11 -| cell mask c5r2.south);
\coordinate (C5) at (C11 -| cell mask c6r2.south);

\path
  let \p1 = ($(ext cell bl)-(ext cell br)$),
      \n1 = {veclen(\x1,\y1)} in
  node[text width=\n1] at ({$(ext cell bl)!.5!(ext cell br)$} |- C1) {%
    \subcaption{Segmented solar cells}%
    \label{fig:el-pv-seg-segmented-cells}%
};

\path
  let \p1 = ($(avg cell bl)-(avg cell br)$),
      \n1 = {veclen(\x1,\y1)} in
  node[text width=\n1] at ({$(avg cell bl)!.5!(avg cell br)$} |- C2) {%
    \subcaption{Average solar cell and its profiles}%
    \label{fig:el-pv-seg-average-cell}%
};

\path
  let \p1 = ($(aug cell bl)-(aug cell br)$),
      \n1 = {veclen(\x1,\y1)} in
  node[text width=\n1] at ({$(aug cell bl)!.5!(aug cell br)$} |- C3) {%
    \subcaption{Augmented mask}%
    \label{fig:el-pv-seg-augmented-mask}%
};

\path
  let \p1 = ($(extra geom bl)-(extra geom br)$),
      \n1 = {veclen(\x1,\y1)} in
  node[text width=\n1] at ({$(extra geom bl)!.5!(extra geom br)$} |- C3) {%
    \subcaption{Extra geometry}%
    \label{fig:el-pv-seg-extra-geom}%
};

\path
  let \p1 = ($(final cell bl)-(final cell br)$),
      \n1 = {veclen(\x1,\y1)} in
  node[text width=\n1] at ({$(final cell bl)!.5!(final cell br)$} |- C3) {%
    \subcaption{Final mask}%
    \label{fig:el-pv-seg-final-mask}%
};

\end{scope}

\node[anchor=east,inner sep=0pt] at (cell mask c2r1.east -| final cell br)
{%
  \pgfplotslegendfromname{cell profiles}%
};

\end{tikzpicture}%

%% file: figures/exact-vs-quad-masks.tex
\begin{tikzpicture}
  [
    subfigures,
    tight background,
  ]
  \graphicspath{{./images/masks/}}
  \matrix
  [
    inner sep=0pt,
    matrix of nodes,
    column sep={1em,between borders},
  ]
  {
    \subcaptionbox{\label{fig:em}}{%
      \input{figures/masks/1_2_em}%
    }%
    &
    \subcaptionbox{\label{fig:qm}}{%
      \input{figures/masks/1_2_qm}%
    }%
    \\
  };
\end{tikzpicture}%

%% file: figures/qualitative-results.tex
\tikzpicturedependsonfile{figures/qualitative/3.tex}%
\tikzpicturedependsonfile{figures/qualitative/Result_of_89457_H.tex}%
\tikzpicturedependsonfile{figures/qualitative/161116_AQUAM_i6300mA_t5000ms_Yingli_04119_PM055852.tex}%
\tikzpicturedependsonfile{figures/qualitative/1_1.tex}%
\pgfmathsetlengthmacro\sep{1em}%
\begin{tikzpicture}
 [
   tight background,
 ]
   \graphicspath{{./images/qualitative/}}%
   \pgfmathsetlengthmacro\side{(\textwidth-\sep*1)/2}

   \matrix
   [
     outer sep=0pt,
     inner sep=0pt,
     column sep={\sep,between borders},
     row sep={\sep,between borders},
     matrix of nodes,
     row 1 column 1/.style={
       execute at begin cell={%
         \setkeys{Gin}{width=\side}%
       },
     },
     row 2 column 1/.style={
       execute at begin cell={%
         \setkeys{Gin}{height=\side}%
       },
     },
     row 1/.style={
       anchor=south,
     },
     row 2/.style={
       anchor=north,
     },
     column 2/.style={
       execute at begin cell={%
         \setkeys{Gin}{width=\side}%
       },
     },
   ] (qualitative)
   {
     \input{figures/qualitative/Result_of_89457_H}%
     &
     \input{figures/qualitative/161116_AQUAM_i6300mA_t5000ms_Yingli_04119_PM055852}%
     \\
     \input{figures/qualitative/3}%
     &
     \input{figures/qualitative/1_1}%
     \\
   };

   \begin{scope}
     [
       inner sep=0pt,
       outer sep=0pt,
       below,
       text width=(\linewidth-\sep)/2,
     ]

     \node at (qualitative-2-1.south) {%
       \subcaption{Monocrystalline}%
     };

     \node at (qualitative-2-2.south) {%
       \subcaption{Polycrystalline}%
     };

   \end{scope}

   \matrix
   [
      /pgfplots/every axis legend,
      matrix of nodes,
      anchor=north,
   ]
   at (qualitative-1-1.north |- qualitative-1-2.north west) {
     \cellmarker
     &
     Solar cells
     &
     \busbarmarker
     &
     Busbars
     \\
   };
\end{tikzpicture}%